\documentclass[11pt,a4paper]{article}

\usepackage[margin=1in]{geometry}
\usepackage[utf8]{inputenc} 
\usepackage[T1]{fontenc}    
\usepackage{hyperref}       
\usepackage{url}            
\usepackage{booktabs}       
\usepackage{amsfonts}       
\usepackage{nicefrac}       
\usepackage{microtype}      
\usepackage{xcolor}         
\usepackage{soul}

\usepackage{subcaption}
\usepackage{hyperref}
\usepackage{url}
\usepackage{mathtools}
\usepackage{amsmath, amssymb, amsfonts, bm, amsthm, bbm,xfrac}
\usepackage{thmtools,thm-restate}
\usepackage{xcolor}
\usepackage{cleveref}
\usepackage{enumitem}
\usepackage{authblk}
\usepackage[round,authoryear]{natbib}

\hypersetup{
    colorlinks = true,
    linkcolor = {blue},
    citecolor = {blue},
    urlcolor = {blue}
}

\numberwithin{equation}{section}
\theoremstyle{definition}

\newtheorem{proposition}{Proposition}

\newtheorem{remark}{Remark}

\newcommand{\expit}{\text{expit}}
\newcommand{\E}{\mathbb{E}}

\title{Annealing in variational inference mitigates mode collapse: A theoretical study on Gaussian mixtures}

\author{%
  Luigi Fogliani$^{1,2}$,\quad
  Bruno Loureiro$^{1}$, \quad
  Marylou Gabrié$^{2}$ \\
  {\small $^{1}$Département d’Informatique, École Normale Supérieure, Université PSL, CNRS}\\
  {\small $^{2}$Laboratoire de Physique de l’École Normale Supérieure, Université PSL, CNRS}\\

}

\begin{document}

\maketitle

\begin{abstract}
Mode collapse --- the failure to capture one or more modes when targeting a multimodal distribution --- is a central challenge in modern variational inference. In this work, we provide a mathematical analysis of annealing-based strategies for mitigating mode collapse in a tractable setting: learning a Gaussian mixture, where mode collapse is known to arise. Leveraging a low-dimensional summary statistics description, we precisely characterize the interplay between the initial temperature and the annealing rate, and derive a sharp formula for the probability of mode collapse. Our analysis shows that an appropriately chosen annealing scheme can robustly prevent mode collapse. Finally, we present numerical evidence that these theoretical trade-offs qualitatively extend to neural network–based models, 
RealNVP normalizing flows, providing guidance for designing annealing strategies mitigating mode collapse in practical variational inference pipelines. 
\end{abstract}

\section{Introduction} 




Variational inference is a widely used framework for approximating complex probability distributions in settings where direct sampling is computationally prohibitive, with applications ranging from scalable Bayesian inference in machine learning to the sampling of Boltzmann distributions and free-energy landscapes in physics, chemistry, and materials science. Given a target distribution $\pi$, variational inference (VI) replaces the task of sampling from $\pi$ with that of finding, within a family of candidate distributions $\mathcal Q$, the probabilistic model $q$ that best approximates $\pi$. The variational family $\mathcal Q$ is restricted to distributions that are easy to sample from and whose probability density function, also denoted by $q$, is known, normalization included. These requirements make the reverse Kullback–Leibler (KL) divergence 
\begin{align}
\label{eq:intro:reverseKL}
    \mathcal D_{\rm KL}(q|\pi)=\mathbb E_q\left[\log\frac{q(x)}{\pi(x)}\right]
\end{align}
a tractable criterion for selecting $q$, and allow samples from $q$ to be used a posteriori as proxies for samples from $\pi$. Notably, optimizing $\mathcal D_{\rm KL}(q|\pi)$ with respect to $q$ does not require having knowledge of the normalization of $\pi$. Rooted in statistical mechanics \citep{weissHypotheseChampMoleculaire1907}, VI was later popularized in Bayesian statistics \citep{jordanIntroductionVariationalMethods1999,bleiVariationalInferenceReview2017}, in particular for its good scaling properties. Moreover, advances in deep generative modelling have greatly expanded the class of tractable variational distributions, with normalizing flows providing flexible parameterizations that retain both easy sampling and tractable probability densities \citep{rezendeVariationalInferenceNormalizing2015,albergo_flow-based_2019}  and autoregressive networks providing the equivalent for discrete variables \citep{wu_solving_2019}.

However, a class of target distributions poses a significant challenge to VI, namely multimodal distributions. The reverse KL objective is well known to be mode seeking \citep{jordanIntroductionVariationalMethods1999,minkaDivergenceMeasuresMessage2005}, yielding approximations that concentrate on a subset of the target modes, a phenomenon commonly referred to as mode collapse. Increasing the expressivity of variational families through deep generative models does not, in general, alleviate this issue \citep{noeBoltzmannGeneratorsSampling2019,hackettFlowbasedSamplingMultimodal2021,nicoliDetectingMitigatingModecollapse2023,blessing_beyond_2024,soletskyi_theoretical_2024,greniouxImprovingEvaluationSamplers2025b}. 

Various empirical strategies have been proposed to mitigate mode collapse in VI in the context of different fields of applications including molecular simulations \citep{noeBoltzmannGeneratorsSampling2019}, classical and quantum field theories \citep{wu_solving_2019,vaitlGradientsShouldStay2022a,nicoliDetectingMitigatingModecollapse2023} and Bayesian inference \citep{wangMitigatingModeCollapse2025}. Among them, a widely applicable approach when optimization of the reverse KL is performed iteratively, typically using stochastic gradient descent, is to combine VI with annealing. Rather than targeting $\pi$ directly, the variational distribution $q$ is first optimized to approximate a tempered distribution proportional to $\pi^\beta$ with $\beta<1$, for which modes are typically connected. The temperature is then progressively lowered during optimization until $\beta=1$ is reached. The early phases of optimization at high ``temperature'' $1/\beta$ promote exploration and mitigate mode collapse. Annealed VI has been successfully applied to sampling from Boltzmann distributions in statistical mechanics \citep{wu_solving_2019,schopmansTemperatureAnnealedBoltzmannGenerators2025}, to sampling from Bayesian posteriors \citep{huangImprovingExplorabilityVariational2018,wangMitigatingModeCollapse2025}, and to optimization tasks, including locating multiple modes \citep{leminhNaturalVariationalAnnealing2025} or identifying global minima \citep{hibat-allah_variational_2021,khandokerLatticeProteinFolding2025}.

Despite the major challenge posed by mode-collapse and the practical successes that annealing has brought, little theoretical work has focused on gaining understanding around its behavior. 
In this work, we make a first step in closing this gap by investigating annealing in a tractable multimodal VI setting. The target is a bimodal Gaussian mixture with well-specified variational family whose parameters are trained by gradient flow. Our \textbf{main contributions} are:
\begin{itemize}
    \item First, we show that mode collapse can be systematically avoided in this setting by employing a suitable annealing schedule, including in parameter regimes where collapse is otherwise unavoidable. These experiments reveal a critical trade-off between the initial temperature and the annealing rate, which must be carefully balanced to reliably recover all target modes.
    \item We show that in the high-dimensional limit the dynamics of the student variances decouple from those of the means and admit a closed-form solution: the variances track the temperature with a delay set by the annealing rate.
    Combined with a set of summary statistics for the means, this recasts the high-dimensional gradient flow into a low-dimensional dynamical system, allowing us to tractably identify distinct high- and low-temperature dynamical regimes and derive a closed-form expression for the mode collapse probability that quantitatively captures the aforementioned experimental trade-off.
    \item We show that the previous findings extend to a practical and expressive choice of variational family, RealNVP normalizing flows \citep{dinh_density_2017}. This suggests that our theoretical results can provide guidance for the design of annealing schedules in practical pipelines.
\end{itemize}

\paragraph{Related works.}
Theoretical investigation of mode collapse in VI has received little attention to this day. By studying the gradient flow associated with VI over Gaussian mixtures with fixed weights, \cite{huix_theoretical_2024} obtained theoretical guarantees in a setting related to the one considered below. Their analysis, however, was not concerned with multimodal target distributions and did not investigate the mode-collapse phenomenon. \cite{talamon_variational_2025} investigated variational inference with Gaussian mixture variational families through gradient-flow-based algorithms, emphasizing computational and algorithmic aspects, while leaving the analysis of mode collapse unexamined. An exception is \citep{soletskyi_theoretical_2024}, from which the present paper draws inspiration.

Building on ideas from statistical physics of learning \citep{saad1995dynamics}, \cite{soletskyi_theoretical_2024} derived a summary statistics description for VI on a bimodal Gaussian mixture. A key difference with respect to our work is that in \citep{soletskyi_theoretical_2024} the variance is kept fixed, making the description not amenable to the study of annealing. Further, \cite{soletskyi_theoretical_2024}  showed that the probability of occurrence of mode-collapse is controlled by the distance between modes of the target mixture with a critical radius $R_c$, function of the class imbalance, above which the dynamics from an uninformed initialization is highly likely to collapse on one mode. 
Using a similar setup, keeping the weights fixed for simplicity, the present work extends the analysis to learnable variances whose dynamics we characterize exactly in the high-dimensional limit, a key ingredient to study annealing which changes the effective variance of the target model along the training dynamics.


Finally, annealing was previously identified as a mean to avoid collapse in VI powered by modern generative models by several authors as already cited in the introduction \citep{wu_solving_2019,schopmansTemperatureAnnealedBoltzmannGenerators2025,huangImprovingExplorabilityVariational2018,wangMitigatingModeCollapse2025,leminhNaturalVariationalAnnealing2025,hibat-allah_variational_2021,khandokerLatticeProteinFolding2025}. Used as a heuristic method, little guidance is provided in these works on how to adjust the annealing schedule. An exception is \citep{wangMitigatingModeCollapse2025}, which proposes an adaptive schedule for the annealing speed based on the instantaneous effective sample size of proposals from a normalizing flow along its training by VI. Comparatively, the adjustment of annealing methods has been extensively discussed in the context of Monte Carlo samplers such as parallel tempering \citep{syedParallelTemperingOptimized2021} or sequential Monte Carlo \citep{barzegarOptimalSchedulesAnnealing2024} and optimizers with simulated annealing \citep{HajekCooling1988,nadlerDynamicsOptimalNumber2007}.
In a complementary direction, \citep{ichikawa2024learning, ichikawa2025high} theoretically analyze the effects of KL annealing in variational autoencoders, deriving sharp thresholds for posterior collapse and characterizing the resulting learning dynamics.
\section{Setup and preliminary experiments}

\subsection{Annealed VI with bimodal Gaussian mixtures}
We consider a minimal multimodal model for mode collapse in VI, where both $\pi$ and $q$ are mixtures of two isotropic Gaussian distributions on $\mathbb{R}^d$:
\begin{align}
    \pi(x) =& w_* \mathcal{N}(x|\mu_*, I_d) + (1-w_*)\mathcal{N}(x|-\mu_*, I_d) ,
    \label{eq:target} \\
    q_\theta(x) =&  w_1 \mathcal{N}(x|\mu_1, \sigma_1^2I_d) + (1-w_1)\mathcal{N}(x|\mu_2, \sigma_2^2I_d), \notag
\end{align}
where $w_*, w_1 \in [0,1]$, $\mu_*, \mu_1, \mu_2 \in \mathbb{R}^d$ and $\sigma_1, \sigma_2 \in \mathbb{R_+}$. Here the trainable parameters of the variational distribution $q_\theta$ are the two means and the two standard deviations: $\theta = (\mu_1, \mu_2, \sigma_1, \sigma_2) \in \mathbb{R}^{2(d+1)}$. 
To simplify the analytical treatment, we fix the weights and choose $w_1=w_{2}=0.5$ in order not to break the symmetry between the modes of the variational model. Following \citep{soletskyi_theoretical_2024} we constrain the means to lie on a sphere $\mu_{\star},\mu_{1,2}\in\mathbb{S}^{d-1}(R)$, which further simplifies the analytical discussion without affecting the qualitative behavior of the model. In this setup, $q_\theta$ is referred to as the student distribution. 

As discussed in the introduction, given the variational family $q_{\theta}$, annealed-VI consists of learning $\theta$ by minimizing the reverse Kullback-Leibler divergence loss between $q_\theta$ and the tempered distribution proportional to $\pi^\beta$ where $1>\beta>0$ is the inverse temperature. Denoting by $\mathcal{Z}_\beta$ the normalization constant of the tempered distribution, the loss is the free energy:
\begin{align}
    \mathcal{L}(\theta, \beta) & = \beta^{-1} \, \mathcal{D}_{\rm KL}\left(q_\theta \, \left\Vert  \,\frac{\pi^\beta}{\mathcal{Z_\beta}}\right.\right) \notag\\
    &= \beta^{-1} \E_{q_\theta} [ \log q_\theta(x)] - w_1 \E_{\mathcal{N}(\mu_1, \sigma_1^2I_d)} [ \log \pi(x)] \notag \\
    & \qquad - w_2 \E_{\mathcal{N}(\mu_2, \sigma_2^2I_d)} [ \log \pi(x)] + \beta^{-1} \log \mathcal{Z}_\beta.
    \label{loss function gmm}
\end{align}
The first term in \cref{loss function gmm} is the entropy of $q_\theta$, which favors exploration pushing $\mu_1$ and $\mu_2$ apart. The two following terms are the cross-entropy between $q_\theta$ and $\pi$, each of them attracting $\mu_{1}$ and $\mu_{2}$, separately, at $\pm \mu_*$. The last term is a constant with respect to $\theta$ and is omitted in the following. The parameter $\beta$ tunes the relative importance between student entropy and cross-entropy with the target, higher temperatures (i.e. lower $\beta$) encouraging exploration. Multiplying the KL divergence between $q_\theta$ and $\pi$ by $\beta^{-1}$ ensures that the derivative of the loss stays of order one throughout annealing.

In practice, $\mathcal{L}(\theta, \beta)$ is minimized with a descent-based algorithm. For simplicity, here we will focus on the spherical gradient flow:
\begin{align}
\label{eq:GF}
\begin{aligned}
        \dot{\mu}_c(t) &= - \nabla_{\mu_c}^{\mathbb{S}^{d-1}(R)}\mathcal{L}\left(\theta(t), \beta(t)\right),\\
    \dot{\sigma}_c(t) &= - \frac{1}{d} \nabla_{\sigma_c}\mathcal{L}\left(\theta(t),\beta(t)\right), \qquad c \in \{1,2\},
\end{aligned}
\end{align}
from an initial $\mu_{1,2}$ uniformly sampled from $\mathbb{S}^{d-1}(R)$ and equal initial variance $\sigma_1^2(0) = \sigma_2^2(0) = \beta^{-1}(0)$. $\beta(t)$ is progressively annealed throughout the trajectory from an initial high-temperature ($\beta < 1$) to the target temperature ($\beta=1$). In particular, we employ the JKO scheme \citep{JKO_1998} on a geometry adapted for isotropic Gaussian mixtures proposed by \cite{talamon_variational_2025}, which allows us to derive the $1/d$ factor difference between the gradient flow equations for the means and for the standard deviations (see Appendix \ref{app:JKO scheme for isotropic gaussians}). This scaling ensures that time derivatives with respect to means and variances stay of order one as $d$ grows.

\begin{figure*}[htb]
\vskip 0.1in
\begin{center}
    \centerline{\includegraphics[width=\linewidth]{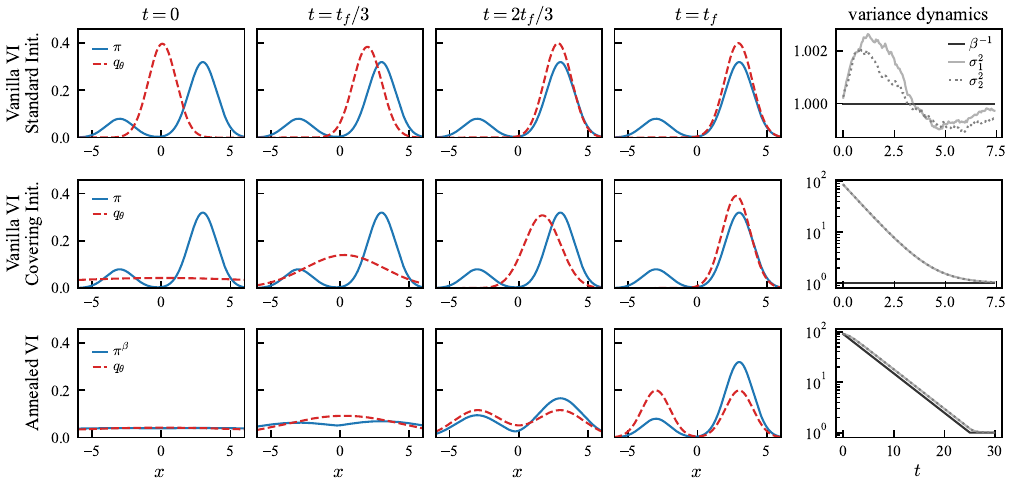}}
    \caption{\textit{Preliminary experiments: annealing mitigates mode collapse.} The first 4 columns show the marginal density along the direction of $\mu_*$ of both $\pi$ and $q_\theta$, at 4 different training stages. The fifth column shows the variance dynamics. Each row represents a different scenario; from top to bottom: no annealing with $\sigma_{1,2}$ initialized to one, no annealing with covering initial distribution (high initial variances $\sigma_{1,2} \gg R$), annealing with schedule given by \cref{exponential annealing schedule}. Hyperparameters values are: $d=512, R=3, w_*=0.8, w_1=0.5$ and time discretization step $\eta=0.05$.
    }
    \label{fig:preliminary experiments}
\end{center}
\vskip -0.2in
\end{figure*}

As shown by \cite{soletskyi_theoretical_2024}, despite its simplicity this setting displays a sufficiently rich phenomenology to investigate mode collapse, and offers the perfect playground to understand how annealing can mitigate it. As we will discuss in the following, our results considerably extend the scope of this work by considering trainable variances for the student. This ingredient is crucial for investigating annealing, as the annealing temperature directly acts on the effective variance of the tempered target $\pi^\beta$. More generally, it allows the student to have a wider distribution, which is believed to be important to avoid mode collapse in VI.

As a starting point, we consider a set of preliminary numerical experiments that will provide us guidance and illustrate interesting phenomenology for the discussion that follows.

\subsection{Preliminary experiments}
\label{sec:preliminary}
We consider a set up where the probability of mode collapse is high with vanilla VI ($\beta=1$ fixed), setting the parameters of $\pi$ to $R=3$ (greater than the $R_c$, the onset radius for mode collapse as characterized in \cite{soletskyi_theoretical_2024}), $w_*=0.8$ and $w_1=0.5$ in dimension $d=512$.
\Cref{fig:preliminary experiments} illustrates the learning dynamics of $q_\theta$ with three different learning strategies. 

Vanilla VI with an initial variance equal to $1$ leads to a collapse of $\mu_1$ and $\mu_2$ to the same mode of $\pi$ (first row). Furthermore, the variances $\sigma_{1}$ and $\sigma_2$ remain close to $1$ throughout the descent, suggesting that the learning dynamics are not very different from the fixed variance setup and, effectively, that trainable variances alone are not sufficient to avoid mode collapse.

A natural idea to avoid mode collapse is to improve the initialization of the gradient descent algorithm by increasing the initial variance of the modes of $q_\theta$ to make it overlap with the two modes of $\pi$. This can be done by setting the initial variances $\sigma_{1,2}^2=\Omega(R^2)$ (second row of \cref{fig:preliminary experiments}). 
Yet, the two modes still collapse to the same mode, and variances rapidly converge to $1$. Moreover, when fixing the values of $\sigma_{1,2}$ to their initial large value, we still observe mode collapse on the same mode (see Appendix \cref{app: fig: supplementary preliminary experiments} first row) and interestingly the variances converge to the target value of $1$ even if we freeze the means (see the second row of \cref{app: fig: supplementary preliminary experiments} in the Appendix).

Finally, we find instead that the learning dynamics with an annealing schedule of the form \footnote{In the experiments with Gaussian mixture students, the gradient flow is discretized with step $\eta=0.05$, and times (including $t_0$) are expressed in gradient-flow units, i.e. $t = \eta \times$ (number of iterations). In the experiments using Adam (\cref{sec:numerical experiments} and Appendix~\ref{appendix: learnable weights}), $t_0$ is instead expressed in number of iterations.
}:
\begin{align}
    \beta(t) = \min(\beta_i^{1-t/t_0}, 1),
    \label{exponential annealing schedule}
\end{align}
with $\beta_i^{-1}=10R^2$ and $t_0=25$ (third row of \cref{fig:preliminary experiments}),
avoids mode collapse. Variances are initialized to $1/\beta_i$ and interestingly, stay close to $1/\beta(t)$ throughout the learning dynamics, a behavior that we derive in \cref{sec:summary}. The schedule in \cref{exponential annealing schedule} is a geometric progression in inverse temperature, which is a common baseline in tempering and has the advantage of remaining analytically tractable. This lets us isolate the trade-off between initial temperature and annealing time, which we discuss next.

These observations illustrate the typical phenomenology observed by practitioners performing VI on a multimodal target distribution with a flexible variational family. Despite a covering initialization, mode collapse can still occur and suitable annealing can be a mean to avoid it. However, choosing the right parameters --- namely the initial temperature $\beta_{i}$ and annealing time  $t_{0}$ --- is crucial to successfully avoid mode collapse. Next, we look closer at this dependency.

\paragraph{Influence of annealing schedule.}
While the efficacy of temperature annealing is well-documented in practice, the selection of an effective schedule remains a non-trivial challenge. To investigate this, we evaluate the impact of the annealing schedule in our controlled setting with an exponential form for $\beta(t)$ given in \cref{exponential annealing schedule}. This choice is motivated by its parameter parsimony and the analytical findings that will follow, which suggest that preventing mode collapse requires the dynamics to remain in a high-temperature regime (specifically, $\beta < 1/R^2$) for a sufficient duration.\footnote{For instance, annealing schemes of the form $\beta(t) = \beta_i + (1-\beta_i)(1-e^{-t/t_0})$ comparatively stay longer in the low temperature regime, which does not help avoiding mode collapse.}

\begin{figure} 
    \centering
    \begin{subfigure}{0.58\linewidth}
        \centering
        \includegraphics[width=\linewidth]{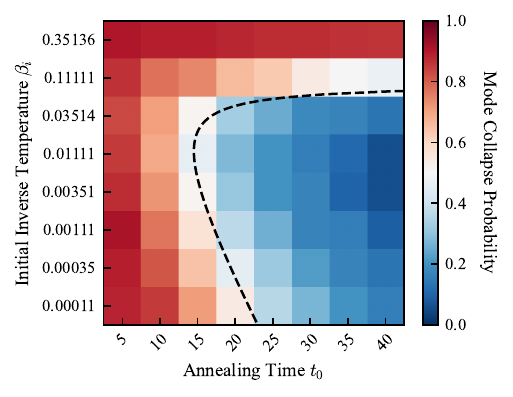}
    \end{subfigure}
    \hfill
    \begin{subfigure}{0.41\linewidth}
        \centering
        \includegraphics[width=\linewidth]{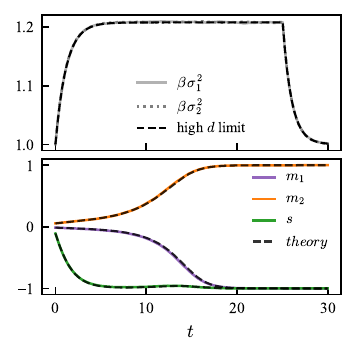}
    \end{subfigure}
    \vspace{-0.5cm}
    \caption{\textbf{(Left)} Annealed VI mode collapse probability with Gaussian mixture student. The annealing scheme is fixed by \cref{exponential annealing schedule} and different values of the initial inverse temperature $\beta_i$ and annealing time $t_0$ are explored. Hyperparameters are the same as in \cref{fig:preliminary experiments}. For each $(\beta_i, t_0)$, the mode collapse probability is computed using 100 different random student mean initializations. The dashed black line is the 0.5 mode collapse isoprobability line obtained with the analytical estimate of \cref{eq: prob collapse}. $\kappa<1$ along this curve.\\
    \textbf{(Top Right)} Dynamics of $\beta(t) \sigma^2(t)$. Grey curves correspond to SGD trajectories of the experiment corresponding to the third row of \cref{fig:preliminary experiments}. The black curve is the closed-form solution \cref{eq: solution of high dimensional limit of variance}. \textbf{(Bottom Right)} Dynamics of the summary statistics for the same experiment. Coloured curves correspond to SGD trajectories and black curves to the numerical integration of the \cref{eq: dynamical system annealing}.
    \label{fig:bimodal grid collapse prob and agreement}}
\end{figure}

\Cref{fig:bimodal grid collapse prob and agreement} (left) reports the probability of mode-collapse over a grid of annealing times ($t_0$) and initial inverse temperatures ($\beta_i$) for the same experimental set up as \cref{fig:preliminary experiments}.
We observe a trade-off: if $\beta_i$ is too high (low initial temperature), the system crosses the $1/R^2$ threshold too rapidly to resolve the target distribution's modes, leading to irreversible collapse. Conversely, while a very low $\beta_i$ provides better exploration, it necessitates a significantly longer annealing time to avoid mode collapse leading to a higher computational budget. This implies the existence of an optimal $\beta_i$ that balances modal resolution with convergence speed. Note that the same qualitative behavior holds with learnable student weights $w_1$, illustrated by \cref{fig: learnable weights prob collapse} in appendix \ref{appendix: learnable weights}.

Our analytical analysis, presented next, leads to an estimation of the mode collapse probability consistent with these empirical observations (0.5 isoprobability line predicted by theory represented by the black dashed line on left panel of \cref{fig:bimodal grid collapse prob and agreement}). Further below, we also show that these observations hold qualitatively for more complex variational families, namely RealNVPs.

\section{Analytical analysis}
Our goal in this section is to give mathematical grounding to the numerical discussion of \cref{sec:preliminary}. Our analysis builds on a low-dimensional reduction of the gradient flow dynamics in \cref{eq:GF}, which we discuss in detail in \cref{sec:summary}. Leveraging this description, we precisely characterize the probability of mode collapse for different choices of initial temperature and annealing rate, showing that a well-tempered annealing schedule is able to successfully mitigate mode collapse.  

\subsection{Summary statistics learning dynamics}
\label{sec:summary}
Consider the gradient flow dynamics on $\mathcal{L}(\theta, \beta)$ introduced in \cref{eq:GF}. Recall that our preliminary numerical experiments (\cref{fig:preliminary experiments}) showed that the student variances track the temperature during annealing, irrespective of the means. The following result, proven in Appendix \ref{app: variance dynamics}, explains this observation.

\begin{proposition}[informal]\label{prop:variance}
In the high-dimensional limit, the derivative of the loss with respect to the variances decouples from the means and from each other:
\begin{align}
    \lim_{d \to \infty} \frac{1}{d} \nabla_{\sigma_c} \mathcal{L}(\theta, \beta) = w_c \, \left( \sigma_c - \frac{1}{\beta\sigma_c}\right), \qquad c \in \{1, 2\} .
\end{align}
\end{proposition}

Consequently, throughout our analysis, we consider the high-dimensional limit of the variance dynamics defined in \cref{eq:GF}:
\begin{equation} \label{eq: high dimensional gf of the variance}
    \dot \sigma_c(t) = -w_c \left(\sigma_c(t) - \frac{1}{\beta(t) \sigma_c(t)} \right), \qquad c\in\{1,2\},
\end{equation}
where the variances are initialized at the same value $\sigma_1^2(0) = \sigma_2^2(0) = \beta_i^{-1}$. Hence, equal weights ($w_1=w_2=0.5$) and equal initialization imply $\sigma_1(t)=\sigma_2(t) := \sigma(t)$ at all times, and $\sigma^2$ relaxes towards $\beta^{-1}$. In particular, for the exponential schedule \cref{exponential annealing schedule}, the above dynamics of \cref{eq: high dimensional gf of the variance} yields
\begin{align}
\label{eq: solution of high dimensional limit of variance}
    \beta(t) \sigma^2(t) = \left\{
    \begin{aligned}
        \frac{1}{1 - \kappa} - \frac{\kappa}{1 - \kappa} e^{-(1-\kappa)t}, \qquad t \leq t_0, \\
        1 + \frac{\kappa(1-e^{-(1-\kappa)t_0})}{1-\kappa}e^{-(t-t_0)}, \qquad t>t_0.
    \end{aligned}
    \right.
\end{align}
where $\kappa$ is the annealing rate, defined as the derivative of $\log \beta(t)$: $\kappa = \log(1/\beta_i)/t_0 >0$. \cref{eq: solution of high dimensional limit of variance} holds for any $\kappa\neq 1$. When $\kappa<1$, $\beta\sigma^2$ reaches a plateau; when $\kappa \geq 1$ (the fastest schedules of \cref{fig:bimodal grid collapse prob and agreement}), the variance cannot follow the temperature and $\beta\sigma^2$ grows. The analysis below assumes $\kappa<1$. 
During the annealing ($t \leq t_0$), after a transient of order one, $\sigma^2(t) \approx \big((1-\kappa)\beta(t)\big)^{-1}$: the variance tracks the temperature with a delay. After the temperature reaches its final value of $1$, $\sigma^2(t)$ goes to one exponentially. The validity of the high-dimensional limit of the variance dynamics is illustrated by the excellent agreement of \cref{eq: solution of high dimensional limit of variance} with the SGD trajectories of the variances in \cref{fig:bimodal grid collapse prob and agreement} (top right) with $d=512$.

We now focus on the trajectory of the means $\mu_{1,2}$ given $\sigma(t)$, which explicitly reads
\begin{align}
    \dot{\mu}_c(t) = -\left( \! I_d \! - \! \frac{\mu_c(t) \mu_c(t)^\top}{R^2} \! \right) \! \nabla_{\mu_c}\mathcal{L}(\theta(t), \beta(t)), \qquad c \in \{1,2\},
    \label{eq: spherical gradient flow}
\end{align}
where the student's means $\mu_c(0)$ are initialized uniformly on the sphere $\mathbb{S}^{d-1}(R)$.

The key idea of our analysis is to note that the reverse-KL loss $\mathcal{L}(\mu_{1,2}, \sigma_{1,2}, \beta)$ can be equivalently described in terms of the following three scalar quantities, akin to \emph{order parameters} in statistical physics
\begin{align}
    m_1 = \frac{\mu_1^{\top} \mu_*}{R^2}, \qquad m_2 = \frac{\mu_2^{\top} \mu_*}{R^2}, \qquad s=\frac{\mu_1^{\top} \mu_2}{R^2}.
    \label{summary statistics}
\end{align}
For $c\in\{1,2\}$, $m_c$ represents the alignment of $\mu_c$ with the target modes $\pm \mu_*$, and $s$ measures the alignment between the two student means. Crucially, the time evolution of these correlations recast the complex $2d$-dimensional flow into a 3-dimensional ordinary differential equation:
\begin{align}\label{eq: dynamical system annealing}
\begin{aligned}
    \dot{m_1} =& - \left[ \frac{(m_2-m_1s)}{\beta(t) \sigma^2(t)}f(s,\sigma(t)) + w_1(1-m_1^2) g(m_1,\sigma(t)) \right], \\
    \dot{m_2} =& - \left[ \frac{(m_1-m_2s)}{\beta(t) \sigma^2(t)} f(s,\sigma(t)) + w_2(1-m_2^2) g(m_2,\sigma(t)) \right], \\
    \dot{s} =& - \left[ \frac{2(1-s^2)}{\beta(t) \sigma^2(t)}f(s,\sigma(t)) + w_1(m_2-m_1s) g(m_1,\sigma(t))  + w_2(m_1-m_2s) g(m_2,\sigma(t)) \right].
\end{aligned}
\end{align}
Detailed computations and expressions for $f$ and $g$ are given in Appendix \ref{app: sec: mean dynamics and dynamical system}. Here, $f(s, \sigma)$ represents the repulsive force arising from the student entropy term, while $g(m, \sigma)$ encapsulates the attractive force from the cross-entropy term. The validity of the gradient flow approximation is confirmed by the agreement of the numerical integration of \cref{eq: dynamical system annealing} with the SGD trajectories on the means (\cref{fig:bimodal grid collapse prob and agreement} (bottom right)).

Notably, the above dynamical system does not depend on $d$, which only plays a role through the typical values of the summary statistics at initialization. In high dimensions, randomly initialized means concentrate near the equator orthogonal to the target axis: $m_c \sim \mathcal{N}(0,1/d)$, with a small overlap $s \sim \mathcal{N}(0,1/d)$. To better understand the learning dynamics described in \cref{eq: dynamical system annealing} and the interplay with the dimension, we analyze two instructive limiting cases of the dynamical system defined by the scaling of the temperature $\beta^{-1}$ with respect to $R^2$.


\subsection{Limiting cases of the dynamics}
The low \textit{temperature regime} ($\beta \gg R^{-2}$) is observed in later stages of the annealing schedule. Recall that at any time, the factor $\beta(t) \sigma^2(t)$ is of order 1, and that after a transient of order $1$, $\sigma^2(t) \approx ((1-\kappa)\beta(t))^{-1}$. Hence, in this regime, $\sigma \ll R$. In Appendix \ref{app: sec: low and high temperature limits}, we show that, in this regime, the cross-entropy term $g$ dominates, while the entropic repulsion $f$ vanishes exponentially:
\begin{align}
f(s,\sigma) &\sim \frac{\sqrt{\pi}}{2} \frac{\sqrt{w_1 w_2}}{\sqrt{1-s}} \frac{\sigma}{R} e^{\frac{- R^2(1-s)}{4 \sigma^2}} \, , \nonumber \\
g(m,\sigma) &= -\tanh{(R^2(m+\epsilon))} + \mathcal{O} \left( \sigma^2 / R^2 \right), \nonumber
\end{align}
where
the bias $\epsilon = \frac{1}{2 R^2}\log \frac{w_*}{1-w_*}$ is an increasing function of $w_*$. Therefore, the evolution of the means according to \cref{eq: dynamical system annealing} becomes independent: $\dot{m_c} = w_c (1-m_c^2)\tanh{(R^2(m_c + \epsilon))}$ for $c \in \{1,2\}$. This creates two distinct basins of attraction for each mean separated by the hyperplane $m = -\epsilon$. The asymptotic fate of each mean is determined solely by whether its value at the onset of this regime is greater or lesser than $-\epsilon$.

A particular case is when no annealing is used ($\beta(t) = 1$, with a sufficiently large $R$), as in \cite{soletskyi_theoretical_2024}, which leads to inevitable mode collapse in high dimensional settings. If $1/\sqrt{d} \ll |\epsilon|$ (as in the preliminary experiments), the two student means will fall within the basin of attraction of the heaviest target mode with very high probability, resulting in mode collapse.

Instead, in the \textit{high temperature} regime ($\beta \ll R^{-2}$), as discussed next, the student means are driven away from each other, and consequently, one of them escapes the basin of attraction of the heaviest mode. Namely, when the temperature is high, $\sigma \gg R$ (because $\beta(t) \sigma^2(t)$ is of order 1) and the entropy term dominates the cross-entropy term:
\begin{align}
    f(s,\sigma) &= w_1 w_2 + \mathcal{O}(R^2/\sigma^2 ) \label{eq: f and g low beta asymptotic} \, , \\
    g(m,\sigma) &= - \sqrt{\frac{2}{\pi}} \frac{R}{\sigma} (m+\epsilon) + \mathcal{O}\left( R^2 /\sigma^2 \right) \nonumber.
\end{align}
Again, see Appendix \ref{app: sec: low and high temperature limits} for a detailed discussion.
Here, the dynamics drive the overlap $s$ to $-1$, therefore effectively projecting the system onto the symmetric subspace where $\mu_1 = -\mu_2$ with both scaling as $1/\sqrt{d}$.

To analyze the stability of this configuration in high dimensions, we linearize the dynamics around the symmetric point $(s=-1, \,m_1 = -m_2 = 0)$. We assume that $\beta(t) \sigma^2(t)$ has reached its plateau $(1-\kappa)^{-1}$. The time evolution of the sum and difference of $m_1$ and $m_2$ decouple from $s$:
\begin{align}
    \dot{m_1} \! + \! \dot{m_2} =& - \left[ 2(1-\kappa)f(-1) + \frac{1}{2} g^\prime(0) \right] (m_1 \! + \! m_2) - g(0), \nonumber \\
    \dot{m_1} \! - \! \dot{m_2} =& - \frac{1}{2} g^\prime(0) (m_1 \! - \! m_2),
    \label{linearized dynamical system}
\end{align}
where the second argument of $f$ and $g$ equal to $((1-\kappa)\beta)^{-1/2}$ has been omitted for clarity. The system exhibits a saddle point structure. From the asymptotics of \cref{eq: f and g low beta asymptotic}, the sum is stable because $2(1-\kappa)f(-1) + \frac{1}{2} g^\prime(0) \approx 2(1-\kappa)w_1w_2 > 0$, however the difference is unstable since $g^\prime(0) \approx - \sqrt{\frac{2}{\pi}} \sqrt{(1-\kappa) \beta} \,R < 0$. The latter instability is the key mechanism to mitigate mode collapse as it drives $m_1$ and $m_2$ away from $0$ and in opposite directions, each of the student mean specializing into a different target mode.

Interestingly, the escape speed from the saddle point $-g^\prime(0) \propto \sqrt{(1-\kappa)\beta} \, R$ is a decreasing function of the temperature and the annealing rate $\kappa$. This implies that arbitrarily high temperatures and annealing rates are comparatively less efficient at driving the specialization of the student means into different modes than temperatures closer (but still superior) to $R^2$. Leveraging this high-temperature dynamics, we derive next an analytical estimate for the probability of mode collapse.

\subsection{Mode collapse probability}

Mode collapse occurs if the student means have not been sufficiently driven away during the high temperature regime to fall into distinct basin of attraction of the low temperature regime. Assuming the transition from the high temperature regime to low temperature regime is sharp and occurs at inverse temperature $\beta=\alpha R^{-2}$ --- where $\alpha$ is a constant of order $1$ --- yields the following mode collapse probability $p(\beta)$ for an annealing schedule $\beta$ and random initialization of the student means:
\begin{align}
    p(\beta) =\int_{-2\epsilon e^{-I(\beta)}}^{2\epsilon e^{-I(\beta)}} {\rm d}x \sqrt{\dfrac{d}{4 \pi}}e^{-dx^2/4},
    \label{eq: prob collapse}
\end{align}
where $ I(\beta) = \int_0^{t_1} ds \sqrt{\frac{R^2 (1-\kappa)\beta(s)}{2 \pi}}$ and $t_1$ is the transition time from the high to low temperature regime, defined implicitly by $\beta(t_1)=\alpha R^{-2}$. Details for the derivation of this estimate are provided in Appendix \ref{app: sec: mode collapse probability estimate}. For the exponential schedule \cref{exponential annealing schedule}, neglecting the initial transient of the variance, $\sigma^2=((1-\kappa)\beta)^{-1}$ and
\begin{align}
    I(\beta_i, t_0) = \sqrt{\frac{2}{\pi}} \frac{\sqrt{1-\kappa}}{\kappa} ( \sqrt{\alpha} - \sqrt{R^2 \beta_i}),
\end{align}
from which we compute a theoretical mode collapse probability that exhibits excellent agreement with the experimental results for $\alpha \approx 0.85$ (see isoprobability line of \cref{fig:bimodal grid collapse prob and agreement} (left) and \cref{app: fig: collapse probability slices} in the Appendix). In the limit of high initial temperature $R^2\beta_i \to 0$, the integral --- and consequently the mode collapse probability --- depends solely on the annealing rate $\kappa = \log(1/\beta_i)/t_0$. Namely, the probability  scales as the inverse of the annealing rate: $I(\beta_i, t_0) \sim \sqrt{\frac{2}{\pi}} \frac{ \sqrt{(1-\kappa)\alpha}}{\kappa}$. This implies that for sufficiently high initial temperatures, the annealing rate emerges as the relevant parameter to determine annealing success. \Cref{fig: summary probability collapse} in Appendix confirms that this asymptotic behavior is already at play in the preliminary experiments: as the initial temperature increases, the curves rapidly converge toward this asymptotic regime where the mode collapse probability is dictated purely by the annealing rate. 

The success of annealing depends on maximizing the integral $I(\beta_i, t_0)$, which has one global maximum
(more details in Appendix \cref{app: sec: prob collapse for exponential schedule}). Important deviation from this optimum leads to failure in two different ways. If the initial temperature is set too low ($\beta_i > \beta_i^*$), the duration of the high-temperature regime is too short to generate sufficient mode separation and prevent collapse. Conversely, for a fixed annealing time $t_0$ , raising the initial temperature ($\beta_i < \beta_i^*$) can end up being detrimental because the integrand --- equal to the escape speed $\sqrt{R^2(1-\kappa) \beta(s)}$ --- remains negligible for a comparatively longer time. Consequently, when the true mode separation $R$ is unknown, the most robust strategy for tuning the annealing schedule is to scale $t_0$ and $\beta_i$ in successive trials in a way that ensures that the annealing rate is not increasing.


\section{Numerical experiments with normalizing flows}
\label{sec:numerical experiments}

\paragraph{Experimental setup.}

In this section, we validate our theoretical findings in a practical setting by learning a multimodal target using a highly expressive variational family: RealNVPs normalizing flows \citep{dinh_density_2017}. While generative models like RealNVPs offer significant flexibility, high expressivity alone does not resolve mode collapse for multimodal targets. Consequently, annealing is an increasingly common strategy for mitigating it in practice.

To isolate the influence of the annealing schedule, we retain the bimodal Gaussian target distribution (Equation \ref{eq:target}) in dimension $d=128$ and employ the exponential annealing schedule defined in \cref{exponential annealing schedule}. The student distribution $q_\theta$ is parameterized using a RealNVP.

A normalizing flow defines a parametric invertible map $\Phi_\theta : \mathbb{R}^d \to \mathbb{R}^d$ with a tractable Jacobian, optimized to transport samples from a base distribution $\rho_0$ to the target $\pi$. In this implementation, we employ a standard Gaussian base distribution $\rho_0(z)=\mathcal{N}(0,I_d)$ in dimension $d=128$. Our RealNVP architecture consists of 8 affine coupling layers. The internal scale and translation networks within each coupling block are multi-layer perceptrons (MLPs) with 4 hidden layers and a hidden dimension of $4d=512$.

The target modes are separated by $R=3$ with the weight imbalance $w_*=0.8$. Using the change of variable formula, the loss is estimated as a Monte Carlo estimator of:
\begin{align}
    \mathcal{L}(\theta, \beta) = \beta^{-1} \int_{\mathbb{R}^d} dz \rho_0(z) \log \left[ \frac{\rho_0(z) \left| \det \frac{\partial \Phi_\theta}{\partial z}(z) \right|^{-1}}{\pi(\Phi_\theta(z))^{\beta}} \right].
\end{align}

Unlike the setting of a Gaussian mixture student, where the overlap $s$ serves as a direct order parameter, detecting mode collapse in high-dimensional flows requires a derived metric. We utilize the variance of the generated samples projected onto $\hat{\mu} = \mu_*/R$. Specifically, we monitor $\mathcal{V} = \text{Var}_{x \sim q_\theta} [x \cdot \hat{\mu}]$. A collapsed model capturing only a single mode (variance $\approx 1$) yields $\mathcal{V} \approx 1$, whereas a model that correctly learns the target mixture yields a significantly higher variance ($\mathcal{V} \approx 2.6$ for our parameters). Accordingly, we define a collapse threshold of $\mathcal{V} < 1.6$.

\begin{figure}[t]
    \centering
    \begin{subfigure}{0.50\linewidth}
        \centering
        \includegraphics[width=\linewidth]{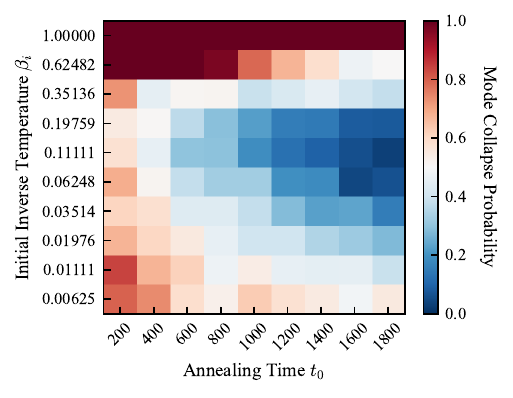}
    \end{subfigure}
    \hfill
    \begin{subfigure}{0.49\linewidth}
        \centering
        \includegraphics[width=\linewidth]{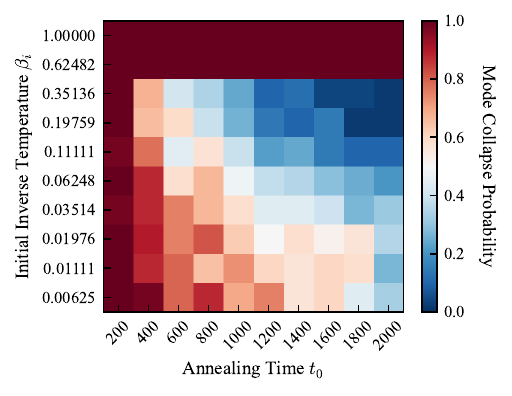}
    \end{subfigure}
    \caption{\textbf{(Left)} Annealed VI mode collapse probability with RealNVP student. We employ the exponential annealing schedule \cref{exponential annealing schedule} on a bimodal Gaussian mixture target with hyperparameters $d=128$, $R=3$ and $w_*=0.8$. For each $(\beta_i, t_0)$ mode collapse probabilities are computed on 200 different runs ($t_0$ in Adam iterations). The generative model is a RealNVP with 8 coupling layers, each containing two MLPs of depth 4 and hidden dimension $4d=512$ (more details in appendix \ref{appendix: realnvp training}). \textbf{(Right)} RealNVP student on a 5-mode target. The target is a Gaussian mixture with 5 isotropic components. All other hyperparameters are the same as in the left panel. Additional details in Appendix~\ref{appendix: 5 mode gaussian target}.}
    \label{fig: realnvp grid prob collapse}
\end{figure}

\paragraph{Results.}
Figure \ref{fig: realnvp grid prob collapse} (left) reports the probability of mode collapse across a grid of initial inverse temperatures $\beta_i$ and annealing times $t_0$ on a bidodal Gaussian mixture target. The results exhibit the same phenomenology observed in the Gaussian mixture student analysis in \cref{fig:bimodal grid collapse prob and agreement} (note that $t_0$ is measured in iterations here). If the initial temperature is too low (high $\beta_i$) or no annealing is applied, mode collapse is inevitable regardless of the model's expressivity. Also, for a fixed computational budget (fixed $t_0$), there exists an optimal initial temperature that maximizes the probability of success; increasing the temperature beyond this point becomes detrimental, as the annealing rate becomes too fast to resolve mode collapse. These results also hold for targets with more modes: \cref{fig: realnvp grid prob collapse} (right) and appendix \ref{appendix: 5 mode gaussian target} show that the same trade-off between initial inverse temperature and annealing time governs collapse on a 5-mode Gaussian mixture, providing evidence that the phenomenology is not specific to bimodal targets (while remaining within the Gaussian-mixture family).

Moreover, \cref{fig: summary probability collapse} in appendix \ref{appendix: dependence on annealing rate} confirms that for sufficiently high initial temperatures, the mode collapse probability depends essentially on the annealing rate alone. This recovers for a complex generative model architecture a behavior predicted by our theoretical analysis, which further confirms the relevance of our theoretical insights in practical settings. 



These results 
are of interest for practitioners dealing with multimodal targets, even when the mode separation $R$ is unknown. To reliably enter the mode collapse mitigation parameter region, one cannot simply increase the initial temperature indefinitely. Instead, our findings indicate that a better strategy is to increase the annealing duration $t_0$ alongside the initial temperature, ensuring that the annealing rate remains sufficiently small to resolve the modes.
\section{Conclusion}

In this work, we have presented a theoretical investigation of annealing in VI on bimodal Gaussian mixture target distribution. Based on the high-dimensional limit of the variance dynamics and a low-dimensional description of the means, we show that simulated annealing can robustly mitigate mode collapse by controlling the balance between the entropy and cross-entropy terms of the reverse Kullback-Leibler divergence. Characterizing the interplay between initial temperature and annealing time, we show that an increase in the initial temperature should be accompanied by an increase in the annealing time when searching for a combination avoiding mode collapse. 
Our numerical experiments show that this strategy works with more practical variational families such as RealNVPs. 

Overall, our theoretical findings, in agreement with our numerical observations, highlight the role of the annealing rate that controls the probability of mode collapse when starting at high enough temperature. Our work considers a geometric annealing scheme, an interesting perspective for future work is the comparison with other schemes such as Effective Sample Size or $\chi^2$-adapted schemes.

Another interesting direction is to replicate this study for VI with diffusion models, a recent promising framework \citep{tzenTheoreticalGuaranteesSampling2019,zhangPathIntegralSampler2021,vargasDenoisingDiffusionSamplers2023a,richterImprovedSamplingLearned2023a,phillipsParticleDenoisingDiffusion2024a,noble_learned_2025}. In this approach, VI is performed on a path measure space which may be more robust to mode collapse.

\section*{Data availability}

A running implementation of the algorithms and all code to reproduce results presented in this work are publicly available at \url{https://github.com/Luijios/annealed_vi_gmms}.

\section*{Acknowledgments}

This work was supported by the PR[AI]RIE-PSAI project, a French government grant managed by the Agence Nationale de la Recherche under the France 2030 program, reference ANR-23-IACL-0008. BL also acknowledges funding from the Choose France - CNRS AI Rising Talents program.

{
\small

\bibliography{references}
\bibliographystyle{abbrvnat}

}


\appendix
\newpage
\appendix

First, we recall the general context of Variational Inference. Given a potentially unnormalized target probability distribution $\pi$ on $\mathbb{R}^d$, and a parametric probability distribution $q_\theta$, the goal is to minimize the reverse Kullback-Leibler objective $\mathcal{D}_{KL}(q_\theta || \pi)$ with respect to $\theta$. In other words, the goal is to find:
\begin{align}
    \theta_* = \underset{\theta \in \Theta}{\text{argmin}} \{\mathcal{D}_{KL}(q_\theta || \pi)\}.
\end{align}
In order to do so, a common strategy is annealing. Introducing an inverse temperature $\beta < 1$, the goal is to relax the objective by minimizing $\mathcal{D}_{KL}(q_\theta || \pi^\beta)$. The inverse temperature is then progressively increased to the target value $\beta=1$.

Here we considered for $q_\theta$ either isotropic Gaussian mixtures, or the probability distribution of a RealNVP normalizing flow. For uniform-weighted Gaussian mixtures,
\begin{align}
    q_\theta = \sum_{i=1}^K \frac{1}{K} \mathcal{N}(\mu_i, \sigma^2_i I_d),
    \label{appendix: isotropic gaussian mixture}
\end{align}
where $\theta = (\mu_1, \dots, \mu_K, \sigma_1, \dots, \sigma_K) \in \mathbb{R}^{K(d+1)}$. In the following section we describe the update rule used to optimize $\theta$.

\section[Appendix]{JKO scheme for isotropic Gaussian mixtures}
\label{app:JKO scheme for isotropic gaussians}

Recall the free energy loss defined in \cref{loss function gmm}:
\begin{align}
    \mathcal{L}(\theta, \beta) & = \beta^{-1} \, \mathcal{D}_{\rm KL}\left(q_\theta \, \left\Vert  \,\frac{\pi^\beta}{\mathcal{Z_\beta}}\right.\right).
\end{align}

In numerical experiments with Gaussian mixtures, we use a JKO scheme \citep{JKO_1998} with a Wasserstein distance adapted to isotropic Gaussians mixtures that has beein introduced by \citep{talamon_variational_2025}. It results in the following discrete update rules, when minimizing the free energy with learning rate $\eta$:
\begin{align}
    \mu_{k+1} &= \mu_k - \eta \beta^{-1} \nabla_{\mu_k} \mathcal{D}_{KL}(q_{\theta_k} || \pi^\beta),\\
    \sigma_{k+1}^2 &= \left(1 - \frac{2\eta \beta^{-1}}{d} \nabla_{\sigma_k^2} \mathcal{D}_{KL}(q_{\theta_k} || \pi^\beta) \right)^2 \sigma_k^2,
\end{align}
where $\mu$ and $\sigma$ respectively denote $\mu_1$ or $\mu_2$, and $\sigma_1$ or $\sigma_2$. This update rule is equivalent to standard gradient descent update on $\mu_k$ and $\sigma_k$ (not $\sigma_k^2$) with a specific scaling between the learning rate of the means and the learning rate of the standard deviations. In fact:
\begin{align}
    \sigma_{k+1} &= \left(1 - \frac{2\eta\beta^{-1}}{d} \nabla_{\sigma_k^2} \mathcal{D}_{KL}(q_\theta || \pi^\beta) \right) \sigma_k \nonumber \\
    &= \sigma_k - \frac{\eta\beta^{-1}}{d} 2 \sigma_k \nabla_{\sigma_k^2} \mathcal{D}_{KL}(q_\theta || \pi^\beta) = \sigma_k - \frac{\eta\beta^{-1}}{d} \nabla_{\sigma_k} \mathcal{D}_{KL}(q_\theta || \pi^\beta).
\end{align}
The JKO scheme implies the $1/d$ factor in front of the derivative with respect to the variances. In terms of the loss the update rule is:
\begin{align}
    \mu_{k+1} &= \mu_k - \eta \nabla_{\mu_k} \mathcal{L}(\theta_k, \beta), \\
    \sigma_{k+1} &= \sigma_k - \frac{\eta}{d} \nabla_{\sigma_k} \mathcal{L}(\theta_k, \beta).
\end{align}

The corresponding gradient flow limit is:
\begin{align}
    \dot \mu(t) &= - \nabla_{\mu} \mathcal{L}(\theta(t), \beta), \\
    \dot \sigma(t) &= - \frac{1}{d} \nabla_{\sigma} \mathcal{L}(\theta(t), \beta),
\end{align}
with $t=\eta k$.

\section{Additional preliminary experiments}

\begin{figure}[h!]
\vskip 0.1in
\begin{center}
    \centerline{\includegraphics[width=\linewidth]{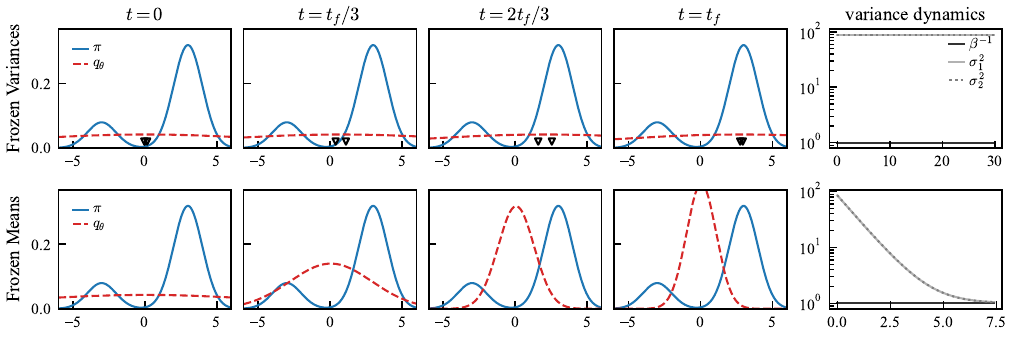}}
    \caption{\textit{Additional preliminary experiments.} The first 4 columns show the marginal density along the direction of $\mu_*$ of both $\pi$ and $q_\theta$, at 4 different training stages. The fifth column shows the variance dynamics. Each row represents a different scenario; from top to bottom: non learned variances with $\sigma_{1,2}^2 =1/10R^2$ fixed, non learned means with $\sigma_{1,2}^2$ initialized to $1/10R^2$. Black triangles on the first row indicate the projection of $\mu_{1,2}$ along the axis of $\mu_*$. Hyperparameters are: $d=512, R=3, w_*=0.8, w_1=0.5$ and time discretization step $\eta=0.05$.}
    \label{app: fig: supplementary preliminary experiments}
\end{center}
\end{figure}

\cref{app: fig: supplementary preliminary experiments} completes \cref{fig:preliminary experiments} and illustrates additional learning strategies. Initial student means are the same as the one of \cref{fig:preliminary experiments}. Frozen variances (first row) lead to mode collapse, even if the variance is set to $1/10R^2$.

Interestingly, the variances rapidly converge to the target value of $1$ even if we freeze the means (second row), in agreement with the high-dimensional decoupling of the variance dynamics (\cref{prop:variance}).

\section{Gradient flow dynamics and low dimensional dynamical system}

In the following, we derive the analytical description of annealed Variational Inference with bimodal Gaussians. By definition, the reverse KL divergence between an isotropic Gaussian mixture with $K$ components and an arbitrary target distribution $\pi$ at inverse temperature $\beta$ is:
\begin{align}
    \mathcal{D}_{KL}(q_\theta || \pi^\beta) = \sum_{i=1}^K \frac{1}{K} \int_{\mathbb{R}^d} dx \frac{e^{-(x-\mu_i)^2/2\sigma_i^2}}{(2 \pi \sigma_i^2)^{d/2}} \log \left( \dfrac{\sum_{j=1}^K \frac{1}{K} \frac{e^{-(x-\mu_j)^2/2\sigma_j^2}}{(2 \pi \sigma_j^2)^{d/2}}}{\pi(x)^\beta} \right).
\end{align}
The goal of this section is to study the gradient flow dynamics related to this loss in the particular case when $q_\theta$ and $\pi$ are bimodal gaussian mixtures.
\begin{align}
    \pi(x) =& w_* \mathcal{N}(x|\mu_*, I_d) + (1-w_*)\mathcal{N}(x|-\mu_*, I_d) \\
    q_\theta(x) =&  w_1 \mathcal{N}(x|\mu_1, \sigma_1^2I_d) + (1-w_1)\mathcal{N}(x|\mu_2, \sigma_2^2I_d)
\end{align}
In what follows, we derive the high-dimensional limit of the variance dynamics (\cref{app: variance dynamics}) and the reduced dynamics of the means (\cref{app: sec: mean dynamics and dynamical system}).

\subsection{Variance dynamics in high dimensions}
\label{app: variance dynamics}

We consider the following gradient flow for the variances, given in \cref{eq:GF} and derived from the JKO scheme (Appendix \ref{app:JKO scheme for isotropic gaussians}):
\begin{align}
    \dot{\sigma}_c(t) = - \frac{1}{d} \nabla_{\sigma_c} \mathcal{L}(\theta(t), \beta(t)).
\end{align}
This dynamics is untractable in general, however, it takes a much simpler form in high dimensions, as shown by the following propositions.

\begin{proposition}
\label{prop: high dimensional limit of the loss}
Assume $\|\mu_1\| = \|\mu_2\| = \|\mu_*\| = R$, with $R$ independent of $d$. Then the loss $\mathcal{L}$ only depends on the variances $\sigma_{1,2}$, the overlap $s = \langle \mu_1, \mu_2 \rangle/R^2$ between the two student means and their respective overlaps $m_{i} = \langle \mu_i, \mu_* \rangle/R^2$ with the teacher mean. In the high-dimensional limit,
    \begin{align}
        \lim_{d \to \infty} \frac{1} d \mathcal{L}(\sigma_1, \sigma_2, s, m_1, m_2, \beta)
        &= \sum_{i \in \{1,2\}} w_i \left[ - \frac{\beta^{-1}}{2} \log (2 \pi e \sigma_i^2) + \frac{\beta^{-1}}{2} \log (2 \pi \beta^{-1}) + \frac{1}{2} \sigma_i^2 \right]
    \end{align}
Moreover, the above convergence is uniform:
    \begin{align}
        \sup \left|\frac1d \mathcal{L}(\sigma_1, \sigma_2, s, m_1, m_2, \beta)
        - \sum_{i \in \{1,2\}} w_i \left[ - \frac{\beta^{-1}}2 \log (2 \pi e \sigma_i^2) + \frac{\beta^{-1}}{2} \log (2 \pi \beta^{-1}) + \frac{1}{2} \sigma_i^2 \right] \right| = O\left(\frac{1}{d}\right),
    \end{align}
where the supremum is taken over $(\sigma_1, \sigma_2, s, m_1, m_2) \in [\sigma_{\min}, \sigma_{\max}]^2 \times [-1, 1]^3$, and $\beta \in [\beta_{\min}, \beta_{\max}]$. $0 < \sigma_{\min} < \sigma_{\max}$ and $0 < \beta_{\min} < \beta_{\max}$ can be arbitrary.
\end{proposition}

\begin{proof}
We consider here the reverse KL divergence loss, where $Z(\beta) = \int_{\mathbb{R}^d} \pi(x)^\beta \, \mathrm{d}x$ normalizes $\pi^\beta$:
\begin{align}
    \mathcal{L}(\sigma_1, \sigma_2, s, m_1, m_2) = \beta^{-1}\E_{Y \sim q_\theta} [\log q_\theta(Y)] - \E_{Y \sim q_\theta} [\log \pi(Y)] + \beta^{-1}\log Z(\beta).
\end{align}
We treat the three terms separately. Denote $k_\sigma(x) = (2\pi\sigma^2)^{-d/2} e^{-\frac{\|x\|^2}{2\sigma^2}}$, so that $q_\theta(y) = \sum_{j \in \{1,2\}} w_j k_{\sigma_j}(y - \mu_j)$.
 
\paragraph{Entropy term.} We have:
\begin{align}
    \E_{Y \sim q_\theta} [\log q_\theta(Y)]
    &= \sum_{i \in \{1,2\}} w_i \E_{\mathcal{N}(\mu_i, \sigma_i^2I_d)} \left[ \log \sum_{j \in \{1,2 \}} w_j k_{\sigma_j} (Y-\mu_j) \right] \\
    &= \sum_{i \in \{1,2\}} w_i \E_{\mathcal{N}(\mu_i, \sigma_i^2I_d)} \left[ \log k_{\sigma_i} (Y-\mu_i) \right] \notag \\
    &\quad + \sum_{i \in \{1,2\}} w_i \E_{\mathcal{N}(\mu_i, \sigma_i^2I_d)} \left[ \log \sum_{j \in \{1,2 \}} w_j \frac{k_{\sigma_j} (Y-\mu_j)}{k_{\sigma_i} (Y-\mu_i)} \right] \\
    &= - \sum_{i \in \{1,2\}} \frac{w_i \, d}{2} \log(2 \pi e \sigma_i^2) + \sum_{i \in \{1,2\}} w_i \E_{\mathcal{N}(\mu_i, \sigma_i^2I_d)} \left[ \log \sum_{j \in \{1,2 \}} w_j \frac{k_{\sigma_j} (Y-\mu_j)}{k_{\sigma_i} (Y-\mu_i)} \right].
\end{align}
The first term is proportional to $d$ and gives a contribution to the high-dimensional limit. The second term is bounded by a constant, and hence vanishes once divided by $d$. Indeed, keeping only the term $j = i$ in the sum gives the lower bound, and Jensen's inequality gives the upper bound:
\begin{align}
     \sum_{i \in \{1,2\}} w_i \log (w_i) &< \sum_{i \in \{1,2\}} w_i \E_{\mathcal{N}(\mu_i, \sigma_i^2I_d)} \left[ \log \sum_{j \in \{1,2 \}} w_j \frac{k_{\sigma_j} (Y-\mu_j)}{k_{\sigma_i} (Y-\mu_i)} \right] \notag \\
    &\leq \sum_{i \in \{1,2\}} w_i \log  \E_{\mathcal{N}(\mu_i, \sigma_i^2I_d)} \left[\sum_{j \in \{1,2 \}} w_j \frac{k_{\sigma_j} (Y-\mu_j)}{k_{\sigma_i} (Y-\mu_i)} \right] \\
    &= \sum_{i \in \{1,2\}} w_i \log \int_{\mathbb{R}^d} k_{\sigma_i}(Y-\mu_i) \sum_{j \in \{1,2 \}} w_j \frac{k_{\sigma_j} (Y-\mu_j)}{k_{\sigma_i} (Y-\mu_i)} \, \mathrm{d}Y \\
    &= 0.
\end{align}
Combining the terms, the high-dimensional limit of the entropy term is
\begin{align}
    \lim_{d \to \infty} \frac{1}{d} \E_{Y \sim q_\theta} [\log q_\theta(Y)] = - \sum_{i \in \{1,2\}} \frac{w_i}{2} \log(2 \pi e \sigma_i^2).
\end{align}
 
\paragraph{Cross-entropy term.} We decompose $y = y^\parallel + y^\bot$, with $y^\parallel$ colinear to $\mu_*$ and $y^\bot$ in the $(d-1)$-dimensional space orthogonal to $\mu_*$, and we identify $y^\parallel$ with its coordinate $\langle y, \mu_* \rangle / R$ along $\mu_*$. We have:
\begin{align}
    \log \pi(y) = - \frac{d}{2} \log(2\pi) - \frac{\|y\|^2}{2} - \frac{R^2}{2} + h(y^\parallel),
\end{align}
where $h(y^\parallel) = \log ( w_* e^{y^\parallel R} + (1-w_*) e^{-y^\parallel R} )$. For $y \sim \mathcal{N}(\mu_i, \sigma_i^2 I_d)$, $y^\parallel$ is a one-dimensional gaussian with variance $\sigma_i^2$ whose law does not depend on $d$, and $h$ grows at most linearly:
\begin{align}
    \log \min(w_*, 1 - w_*) - |z| R \leq h(z) \leq |z| R,
\end{align}
so that $\E [h(y^\parallel)]$ is bounded by a constant independent on $d$ (depending on $\sigma_{\max}, R, w_*$). Therefore,
\begin{align}
    \E_{Y \sim q_\theta} [\log \pi(Y)] = \sum_{i \in \{1,2\}} w_i \E_{\mathcal{N}(\mu_i, \sigma_i^2I_d)} [\log \pi(Y)] = - \frac{d}{2} \log(2 \pi) - \sum_{i \in \{1,2\}} \frac{w_i \, d}{2} \sigma_i^2 + O(1),
\end{align}
and
\begin{align}
    \lim_{d \to \infty} - \frac{1}{d} \E_{Y \sim q_\theta} [\log \pi(Y)] = \sum_{i \in \{1,2\}} \frac{ \, w_i}{2} \left(\log (2 \pi) + \sigma_i^2 \right).
\end{align}
The orthogonal component $y^\bot$ dominates, and the cross-entropy behaves as if $\pi$ were the standard Gaussian.
 
\paragraph{Normalization term.} Using the same decomposition,
\begin{align}
    Z(\beta) &= (2\pi)^{-\frac{\beta d}{2}} \int_{\mathbb{R}^{d-1}} e^{-\beta \frac{\|y^\bot\|^2}{2}} \, \mathrm{d}y^\bot \int_{\mathbb{R}} \left( w_* e^{-\frac{(y^\parallel - R)^2}{2}} + (1-w_*) e^{-\frac{(y^\parallel + R)^2}{2}} \right)^\beta \, \mathrm{d} y^\parallel \\
    &= (2\pi)^{-\frac{\beta d}{2}} \left( \frac{2 \pi}{\beta} \right)^{\frac{d-1}{2}} \int_{\mathbb{R}} \left( w_* e^{-\frac{(y^\parallel - R)^2}{2}} + (1-w_*) e^{-\frac{(y^\parallel + R)^2}{2}} \right)^\beta \, \mathrm{d} y^\parallel.
\end{align}
By the upper bound on $h$, the last integral is finite, positive and independent of $d$. Hence,
\begin{align}
    \lim_{d \to \infty} \frac1d \log Z(\beta) = \frac{1 - \beta}{2} \log(2\pi) - \frac12 \log \beta.
\end{align}
Summing the three contributions gives the result.
\end{proof}

The gradient flow \cref{eq:GF} involves the derivative of the loss with respect to the variances, so we also need the convergence of this derivative. It holds uniformly on compact sets of variances.
 
\begin{proposition}\label{prop:derivative limit}
Let $0 < \sigma_{\min} \leq \sigma_{\max}$. Then, for $c \in \{1,2\}$,
\begin{align}
    \sup \left| \frac1d \frac{\partial \mathcal{L}}{\partial \sigma_c}(\sigma_1, \sigma_2, s, m_1, m_2) - w_c \left( \sigma_c - \frac{1}{\beta \sigma_c} \right) \right| = O \left( \frac{1}{\sqrt{d}} \right),
\end{align}
where the supremum is taken over $\sigma_1, \sigma_2 \in [\sigma_{\min}, \sigma_{\max}]$, all admissible overlaps $s, m_1, m_2$ and all values of $\beta \in [\beta_{\min}, \beta_{\max}]$. Note that $w_c \left( \sigma_c - \frac{1}{\beta \sigma_c} \right)$ is the derivative with respect to $\sigma_c$ of the limit in \Cref{prop: high dimensional limit of the loss}.
\end{proposition}

\begin{proof}
By symmetry, we take $c = 1$. The normalization term $\log Z(\beta)$ does not depend on $\sigma_1$, so we only need to treat the entropy and the cross-entropy terms. Throughout, $\xi \sim \mathcal{N}(0, I_d)$ and $\zeta \sim \mathcal{N}(0, 1)$.
 
\paragraph{Cross-entropy term.} Recall the definition of $h$ in the the proof of \cref{prop: high dimensional limit of the loss}: $h(z) = \log \left( w_* e^{Rz} + (1 - w_*) e^{-Rz} \right)$, so that
\begin{align}
    \log \pi(y) = - \frac{d}{2} \log(2 \pi) - \frac{\|y\|^2}{2} - \frac{R^2}{2} + h(y^\parallel).
\end{align}
Since $\E_{\mathcal{N}(\mu_c, \sigma_c^2 I_d)} \|y\|^2 = R^2 + d \sigma_c^2$ and $y^\parallel \sim \mathcal{N}(m_c / R, \sigma_c^2)$, this gives the exact expression
\begin{align}
    \E_{y \sim q_\theta} [\log \pi(y)] = - \frac{d}{2} \log(2 \pi) - R^2 - \sum_{i \in \{1,2\}} \frac{w_i \, d}{2} \sigma_i^2 + \sum_{i \in \{1,2\}} w_i \, \E \left[ h \left( R m_i + \sigma_i \zeta \right) \right].
\end{align}
Since $|h'| \leq R$, we can differentiate under the expectation, and
\begin{align}
    \left| \frac{\partial}{\partial \sigma_1} \E \left[ h \left( \frac{m_1}{R} + \sigma_1 \zeta \right) \right] \right| = \left| \E \left[ h' \left( \frac{m_1}{R} + \sigma_1 \zeta \right) \zeta \right] \right| \leq R \, \E |\zeta| = R \sqrt{\frac{2}{\pi}}.
\end{align}
Hence, uniformly over all $\sigma_1, \sigma_2 > 0$ and all overlaps,
\begin{align}
    \left| - \frac{1}{d} \frac{\partial}{\partial \sigma_1} \E_{Y \sim q_\theta} [\log \pi(Y)] - w_1 \sigma_1 \right| \leq \frac{R}{\sqrt{2 \pi} \, d}.
\end{align}
 
\paragraph{Entropy term.} We write $\E_{y \sim q_\theta} [\log q_\theta(y)] = w_1 \E [\log q_\theta(\mu_1 + \sigma_1 \xi)] + w_2 \E [\log q_\theta(\mu_2 + \sigma_2 \xi)]$ and differentiate. The terms coming from the explicit dependence of $q_\theta$ on $\sigma_1$ sum to 
\begin{align}
    \E_{y \sim q_\theta} [\partial_{\sigma_1} \log q_\theta(y)] = \partial_{\sigma_1} \int q_\theta = 0,
\end{align}
so that
\begin{align}
    \frac{\partial}{\partial \sigma_1} \E_{y \sim q_\theta} [\log q_\theta(y)] = w_1 \E_{\xi} \left[ \xi \cdot \nabla \log q_\theta(\mu_1 + \sigma_1 \xi) \right].
\end{align}
Denote by
\begin{align}
    p_j(y) = \frac{ w_j \, k_{\sigma_j}(y - \mu_j)}{w_1 \,k_{\sigma_1}(y - \mu_1) + w_2 \, k_{\sigma_2}(y - \mu_2)}
\end{align}
the responsibilities, so that $p_1 + p_2 = 1$ and $\nabla \log q_\theta(y) = - \sum_{j \in \{1,2\}} p_j(y) \frac{y - \mu_j}{\sigma_j^2}$. With $y = \mu_1 + \sigma_1 \xi$ and $\Delta \mu = \mu_1 - \mu_2$, we get
\begin{align}
    \frac{\partial}{\partial \sigma_1} \E_{y \sim q_\theta} [\log q_\theta(y)] = - \frac{w_1d}{\sigma_1} + w_1 \left( \frac{1}{\sigma_1} - \frac{\sigma_1}{\sigma_2^2} \right) \E_\xi \left[ p_2(y) \|\xi\|^2 \right] - \frac{w_1}{\sigma_2^2} \E_\xi \left[ p_2(y) \, \xi \cdot \Delta \mu \right].
\end{align}
The first term is the derivative of $- \frac{w_1d}{2} \log(2 \pi e \sigma_1^2)$, and gives the limit $- \frac{w_1}{\beta \sigma_1}$ once multiplied by $\beta^{-1}$ and divided by $d$. Since $0 \leq p_2(y) \leq 1$ and $\|\Delta \mu\| \leq 2R$, the last term is bounded by $\frac{R}{\sigma_2^2} \sqrt{\frac{2}{\pi}}$, and vanishes once divided by $d$.
 
The second term is of order $d$ a priori: the components cannot be told apart when $\sigma_1 \approx \sigma_2$, but then the prefactor $\frac{1}{\sigma_1} - \frac{\sigma_1}{\sigma_2^2}$ is small. To quantify this, note that $p_2(y) \leq \min \left(1, \frac{w_2 \, k_{\sigma_2}(y - \mu_2)}{w_1 \,k_{\sigma_1}(y - \mu_1)} \right)$, so that, with $\delta = |\sigma_1 - \sigma_2|$ and $S = \sigma_1^2 + \sigma_2^2$,
\begin{align}
    \E_{y \sim \mathcal{N}(\mu_1, \sigma_1^2 I_d)} [p_2(y)]
    &\leq \int_{\mathbb{R}^d} \min \left( k_{\sigma_1}(y - \mu_1), \frac{w_2}{w_1} k_{\sigma_2}(y - \mu_2) \right) \mathrm{d}y
    \leq \sqrt{\frac{w_2}{w_1}} \int_{\mathbb{R}^d} \sqrt{k_{\sigma_1}(y - \mu_1) \, k_{\sigma_2}(y - \mu_2)} \, \mathrm{d}y \notag \\
    &\leq \sqrt{\frac{w_2}{w_1}} \left( \frac{2 \sigma_1 \sigma_2}{S} \right)^{\frac{d}{2}}
    \leq \sqrt{\frac{w_2}{w_1}} e^{- \frac{d \delta^2}{2 S}},
\end{align}
where we computed the Gaussian integral and used $\frac{2 \sigma_1 \sigma_2}{S} = 1 - \frac{\delta^2}{S} \leq e^{- \frac{\delta^2}{S}}$. By the Cauchy--Schwarz inequality and $p_2(y)^2 \leq p_2(y)$, 
\begin{align}
    \E [p_2(y) \|\xi\|^2] \leq \sqrt{\E [p_2]} \sqrt{\E \|\xi\|^4} \leq (d + 1) e^{- \frac{d \delta^2}{4 S}}.
\end{align}
Since $\left| \frac{1}{\sigma_1} - \frac{\sigma_1}{\sigma_2^2} \right| = \frac{\delta (\sigma_1 + \sigma_2)}{\sigma_1 \sigma_2^2}$ and $\sup_{\delta \geq 0} \delta e^{- \frac{d \delta^2}{4 S}} = \sqrt{\frac{2 S}{e d}}$, the second term divided by $d$ is bounded by
\begin{align}
    \frac{d + 1}{2 d} \frac{\sigma_1 + \sigma_2}{\sigma_1 \sigma_2^2} \sqrt{\frac{2 S}{e d}} \leq \frac{4 \sigma_{\max}^2}{\sigma_{\min}^3 \sqrt{e d}}.
\end{align}
 
Combining the three bounds, all of which are uniform over $\sigma_1, \sigma_2 \in [\sigma_{\min}, \sigma_{\max}]$ and all overlaps, gives the result.
\end{proof}
 
\begin{remark}
The rate $d^{-1/2}$ comes from the second term of the entropy derivative, and is only attained when $|\sigma_1 - \sigma_2|$ is of order $d^{-1/2}$. Otherwise, this term vanishes (when $\sigma_1 = \sigma_2$) or is exponentially small in $d$ (when $|\sigma_1 - \sigma_2|$ is bounded away from $0$), and the rate is $d^{-1}$. Note also that the bound on the cross-entropy part is uniform over all $\sigma_1, \sigma_2 > 0$, while the bound on the entropy part uses $\sigma_1, \sigma_2 \in [\sigma_{\min}, \sigma_{\max}]$.
\end{remark}

\subsection{Gradient flow equation for the means}
\label{app: sec: mean dynamics and dynamical system}

In this section we derive the learning dynamics of the means presented in the analytical analysis section of the main body. As a preliminary, we need to compute and simplify $\mathcal{D}_{KL}(q_\theta || \pi^\beta)$ when $q_\theta$ is a Gaussian mixture of the form \cref{appendix: isotropic gaussian mixture}, in the particular case of \textit{equal variances}: $\sigma = \sigma_1 = \dots = \sigma_K$. --- We restrict to this case since we will eventually consider the high dimensional limit with equal initialization. --- The loss simplifies :
\begin{align}
    \beta^{-1}\mathcal{D}_{KL}(q_\theta || \pi^\beta) =& -\beta^{-1}\frac{d}{2} \log(2 \pi e\sigma^2) + \beta^{-1}\int_{\mathbb{R}^d} dx \frac{e^{-x^2/2}}{(2 \pi)^{d/2}} \sum_i \frac{1}{K} \log \left( \sum_{j=1}^K \frac{1}{K} e^{-\frac{x^T(\mu_i - \mu_j)}{\sigma} - \frac{|\mu_i-\mu_j|^2}{2\sigma^2}} \right) \\
    &- \int_{\mathbb{R}^d} dx \frac{e^{-x^2/2}}{(2 \pi)^{d/2}} \sum_i \frac{1}{K} \log (\pi(\sigma x + \mu_i)). \\
\end{align}
Now, we set $K=2$ and $\pi$ as a bimodal isotropic Gaussian mixture and use notations of \cref{eq:target}. We fix the separation radius $|\mu_*|=R$ and the radii of the student means $=|\mu_k|$. This yields:
\begin{align}
    \beta^{-1}\mathcal{D}_{KL}(q_\theta || \pi^\beta) =& - \beta^{-1}\frac{d}{2} \log(2 \pi e\sigma^2) - \beta^{-1}\frac{R^2}{\sigma^2} + \beta^{-1}\int_{\mathbb{R}^d} dx \frac{e^{-x^2/2}}{(2 \pi)^{d/2}} \sum_{i=1}^2 w_i \log \left( \sum_{j=1}^2 w_j e^{\frac{x^T \mu_j}{\sigma} + \frac{\mu_i^T\mu_j}{\sigma^2}} \right) \\
    &+ \frac{d\sigma^2}{2} - \int_{\mathbb{R}^d} dx \frac{e^{-x^2/2}}{(2 \pi)^{d/2}} \sum_{i=1}^2 w_i \log \left( w_* e^{(\sigma x + \mu_i) \mu_*} + (1-w_*) e^{-(\sigma x + \mu_i) \mu_*} \right)
    \label{app: eq: loss}
\end{align}
up to an additional constant that does not depend on $\mu_i$ nor on $\sigma$. Note that the loss only depends on the vectors $\mu_*, \mu_{1,2}$ through the following statistics:
\begin{align}
    m_i = \frac{\mu_i . \mu_x}{R^2}, \quad s = \frac{\mu_1 . \mu_2}{R^2}.
\end{align}
In the following, we alleviate notations by replacing $\int_{\mathbb{R}^d} dx \frac{e^{-x^2/2}}{(2 \pi)^{d/2}}[\dots]$ with $\mathbb{E}_{x \sim \mathcal{N}(0, I_d)}[\dots]$

Let's first compute the dynamics of the means. By definition, the spherical gradient flow is given by:
\begin{align}
    \dot{\mu_c}(t) = \left( I_d - \frac{\mu_c(t) \mu_c(t)^T}{R^2} \right) \nabla_{\mu_c}\mathcal{L}(\theta(t), \beta(t)) \quad c=1,2.
\end{align}
Let's first compute the Euclidean gradient of the KL divergence \cref{app: eq: loss} with respect to $\mu_{1,2}$.
\begin{align}
    \nabla_{\mu_1} \mathcal{D}_{KL} = w_1 \mathbb{E}_{x \sim \mathcal{N}(0, I_d)} \left[ \frac{1}{\sigma} \frac{w_1 (x + 2\frac{\mu_1}{\sigma})e^{(x+\frac{\mu_1}{\sigma}).\frac{\mu_1}{\sigma}} + w_2 \frac{\mu_2}{\sigma}e^{(x+\frac{\mu_1}{\sigma}).\frac{\mu_2}{\sigma}}}{w_1 e^{(x+\frac{\mu_1}{\sigma}).\frac{\mu_1}{\sigma}} + w_2 e^{(x+\frac{\mu_1}{\sigma}).\frac{\mu_2}{\sigma}}}\right] \nonumber\\
    + w_2 \mathbb{E}_{x \sim \mathcal{N}(0, I_d)} \left[ \frac{1}{\sigma} \frac{w_1(x+\frac{\mu_2}{\sigma})e^{(x+\frac{\mu_2}{\sigma}).\frac{\mu_1}{\sigma}}}{w_1e^{(x+\frac{\mu_2}{\sigma}).\frac{\mu_1}{\sigma}} + w_2e^{(x+\frac{\mu_2}{\sigma}).\frac{\mu_2}{\sigma}}} \right] \nonumber\\
    - \beta w_1\mathbb{E}_{x \sim \mathcal{N}(0, I_d)} \left[ \frac{w_* \mu_* e^{(\sigma x + \mu_1).\mu_*} - (1-w_*) \mu_* e^{-(\sigma x + \mu_1).\mu_*}}{w_* e^{(\sigma x + \mu_1).\mu_*} + (1-w_*) e^{-(\sigma x + \mu_1).\mu_*}} \right]
\end{align}
Defining the sigmoid function $\expit(t) = (1+e^{-x})^{-1}$, we can rewrite:
\begin{align}
    \nabla_{\mu_1} \mathcal{D}_{KL} = \frac{w_1}{\sigma} \mathbb{E}_{x \sim \mathcal{N}(0, I_d)} \left[ (x + 2\frac{\mu_1}{\sigma}) \expit \left( (\frac{\mu_1}{\sigma} - \frac{\mu_2}{\sigma}).(x+\frac{\mu_1}{\sigma}) + \log \frac{w_1}{w_2} \right) \right. \nonumber\\
    \left.+ \frac{\mu_2}{\sigma} \expit \left( (\frac{\mu_2}{\sigma} - \frac{\mu_1}{\sigma}).(x+\frac{\mu_1}{\sigma}) + \log \frac{w_2}{w_1} \right) \right] \nonumber \\
    + \frac{w_2}{\sigma} \mathbb{E}_{z \sim \mathcal{N}(0, I_d)} \left[ (x+\frac{\mu_2}{\sigma}) \expit \left( (\frac{\mu_1}{\sigma}-\frac{\mu_2}{\sigma}).(x+\frac{\mu_2}{\sigma}) + \log \frac{w_1}{w_2} \right) \right] \nonumber \\
    + \beta w_1\mathbb{E}_{z \sim \mathcal{N}(0, I_d)} \left[ \mu_* \left(1 -2\expit \left(2\mu_*.(\sigma x+\mu_1) + \log \frac{w_*}{1-w_*} \right) \right) \right]
\end{align}
It's convenient to eliminate $xg(x)$-like terms using Stein's lemma. Since $x \sim \mathcal{N}(0, I_d)$, we have $\mathbb{E}_x[xg(x)] = \mathbb{E}_x [\nabla g(x)]$, and:
\begin{align}
    \nabla_{\mu_1} \mathcal{D}_{KL} = w_1 \mathbb{E}_{x \sim \mathcal{N}(0, I_d)} \left[ \frac{\mu_1}{\sigma^2} \expit \left( (\frac{\mu_1}{\sigma} - \frac{\mu_2}{\sigma}).(x+\frac{\mu_1}{\sigma}) + \log \frac{w_1}{w_2} \right) \left( 3 - \expit \left( (\frac{\mu_1}{\sigma} - \frac{\mu_2}{\sigma}).(x+\frac{\mu_1}{\sigma}) + \log \frac{w_1}{w_2} \right) \right) \right. \nonumber \\
    \left. + \frac{\mu_2}{\sigma^2} \expit \left( (\frac{\mu_2}{\sigma} - \frac{\mu_1}{\sigma}).(x+\frac{\mu_1}{\sigma}) + \log \frac{w_2}{w_1} \right)^2 + \beta \mu_* \left(1-2\expit \left(2\mu_*.(\sigma x+\mu_1)+\log \frac{w_*}{1-w_*}\right) \right) \right] \nonumber \\
    + w_2 \mathbb{E}_{x \sim \mathcal{N}(0, I_d)} \left[ \frac{\mu_1}{\sigma^2} \expit \left( (\frac{\mu_1}{\sigma} - \frac{\mu_2}{\sigma}).(x+\frac{\mu_2}{\sigma}) + \log \frac{w_1}{w_2} \right) \left( 1 - \expit \left( (\frac{\mu_1}{\sigma} - \frac{\mu_2}{\sigma}).(x+\frac{\mu_2}{\sigma}) + \log \frac{w_1}{w_2} \right) \right) \right. \nonumber \\
    \left. + \frac{\mu_2}{\sigma^2} \expit \left( (\frac{\mu_1}{\sigma} - \frac{\mu_2}{\sigma}).(x+\frac{\mu_2}{\sigma})+ \log \frac{w_1}{w_2} \right)^2 \right]
\end{align}
Projecting the gradient with \cref{eq: spherical gradient flow} results in:
\begin{align}
    \nabla_{\mu_1}^{\mathbb{S}} \mathcal{D}_{KL} = \frac{\mu_2 - \mu_1 s}{\sigma^2} \mathbb{E}_{x \sim \mathcal{N}(0, I_d)} \left[ w_1 \expit \left( (\frac{\mu_2}{\sigma} - \frac{\mu_1}{\sigma}).(x+\frac{\mu_1}{\sigma}) + \log \frac{w_2}{w_1}\right)^2 \right. \nonumber \\
    \left. + w_2 \expit \left( (\frac{\mu_1}{\sigma} - \frac{\mu_2}{\sigma}).(x+\frac{\mu_2}{\sigma}) + \log \frac{w_1}{w_2} \right)^2 \right] \nonumber \\
    + \beta w_1 (\mu_* - m_1 \mu_1) \mathbb{E}_{x \sim \mathcal{N}(0, I_d)} \left[1 - 2 \expit \left( 2\mu_*(\sigma x+\mu_1) + \log \frac{w_*}{(1-w_*)} \right) \right].
\end{align}
We notice that the functions inside the expectations depend only on the projections of $x$ in the direction of two axis: $\mu_*.x$ wich has variance $R^2$, and $(\mu_2-\mu_1).x$ wich has variance $2R^2(1-s)$.

Introducing the functions:
\begin{align}
    f(s, \sigma) =& \mathbb{E}_{x \sim \mathcal{N}(0, 1)} \left[ w_1 \expit \left( \frac{R}{\sigma}\sqrt{2(1-s)}x + \frac{R^2}{\sigma^2}(s-1) + \log \frac{w_2}{w_1}\right)^2 + \right. \nonumber \\
    & \left. \hspace{7em} + w_2 \expit \left( \frac{R}{\sigma}\sqrt{2(1-s)}x + \frac{R^2}{\sigma^2}(s-1) + \log \frac{w_1}{w_2}\right)^2\right] \nonumber \\
    g(m,\sigma) &= \mathbb{E}_{x \sim \mathcal{N}(0, 1)} \left[ 1 - 2\expit \left( 2 \sigma Rx + 2R^2m + \log \frac{w_*}{1-w_*} \right) \right]
    \label{app: eq: f and g}
\end{align}
we can conveniently write the gradient flow as:
\begin{align}
    \nabla_{\mu_1}^\mathbb{S} \mathcal{D}_{KL} = \frac{\mu_2 - \mu_1 s}{\sigma^2} f(s, \sigma) + \beta w_1 (\mu_* - m_1 \mu_1) g(m_1, \sigma).
\end{align}
Using the symmetry $\mu_1 \to \mu_2$, $w_1 \to w_2$, we obtain the gradient with respect to $\mu_2$.
\begin{align}
    \nabla_{\mu_2}^\mathbb{S} \mathcal{D}_{KL} = \frac{\mu_1 - \mu_2 s}{\sigma^2} f(s, \sigma) + \beta w_2 (\mu_* - m_2 \mu_2) g(m_2, \sigma).
\end{align}
Finally, we obtain the learning dynamics of the summary statistics by writing:
\begin{align}
    \dot{m_1} &= \frac{\mu_*.\dot\mu_1}{R^2} = - \frac{\mu_*}{R^2} . \beta^{-1} \nabla_{\mu_1}^\mathbb{S} \mathcal{D}_{KL} \nonumber\\
    &= - \left[ \frac{(m_2 - m_1 s)}{\beta\sigma^2} f(s, \sigma) + w_1 (1 - m_1^2) g(m_1, \sigma) \right] \\
    \dot{s} &= - \frac{1}{R^2} [\mu_2 . \beta^{-1}\nabla_{\mu_1}^{\mathbb{S}} \mathcal{D}_{KL} + \mu_1 . \beta^{-1}\nabla_{\mu_2}^{\mathbb{S}} \mathcal{D}_{KL}] \nonumber \\
    &= - \left[ \frac{2(1-s^2)}{\beta \sigma^2} f(s,\sigma) + w_1(m_2-m_1s) g(m_1, \sigma) + w_2(m_1-m_2s) g(m_2,\sigma) \right]
\end{align}

Finally, we get:
\begin{align}
    \dot{m_1} =& - \left[ \frac{(m_2-m_1s)}{\beta(t) \sigma^2(t)}f(s,\sigma(t)) + w_1(1-m_1^2) g(m_1,\sigma(t)) \right] \nonumber \\
    \dot{m_2} =& - \left[ \frac{(m_1-m_2s)}{\beta(t)\sigma^2(t)}f(s,\sigma(t)) + w_2(1-m_2^2) g(m_2,\sigma(t)) \right] \nonumber \\
    \dot{s} =& - \left[ \frac{2(1-s^2)}{\beta(t)\sigma^2(t)}f(s,\sigma(t)) + w_1(m_2-m_1s) g(m_1,\sigma(t)) + w_2(m_1-m_2s) g(m_2,\sigma(t)) \right]
    \label{app: eq: dynamical system annealing}
\end{align}
Numerical integration of these coupled ODEs are in excellent agreement with trajectories obtained by SGD on both means and variances (bottom right panel of \cref{fig:bimodal grid collapse prob and agreement}).

\subsection{Low- and High-variance limits}
\label{app: sec: low and high temperature limits}

Recall that since $\beta(t) \sigma^2(t)$ is of order 1 at any time (\cref{eq: solution of high dimensional limit of variance}), high and low variance limits correspond respectively to high and low temperature limits. Precisely, after an order 1 transient, the variance and the inverse temperature are related by an order one constant: $\beta(t) \sigma^2(t) \approx (1-\kappa)^{-1}$.

\subsubsection{Low variance limit ($\sigma \ll R$)}

Let's begin with $f$. We denote $r = \frac{R}{\sigma}$, $A = \sqrt{2(1-s)}$, $B = s-1 = -(1-s) = -\frac{A^2}{2}$, and $C = \log \left( \frac{w_2}{w_1} \right)$. We consider:
\begin{align}
    &\mathbb{E}_{x \sim \mathcal{N}(0, I_d)} \left[ \expit \left( rAx - r^2A^2/2 + C\right)^2 \right] \nonumber \\
    =&  \mathbb{E}_{x \sim \mathcal{N}(-r^2A^2/2 + C, rA)} \left[ \expit \left( x\right)^2 \right] \nonumber \\
    =& \int_\mathbb{R} \frac{dx}{\sqrt{2 \pi} rA} e^{-\frac{(x + r^2A^2/2 - C)^2}{2r^2A^2}} \expit(x)^2 \nonumber \\
    =& \frac{1}{\sqrt{2 \pi} rA} e^{-\frac{(r^2A^2/2 - C)^2}{2 r^2 A^2}} \int_\mathbb{R} dx e^{-\frac{x^2}{2r^2A^2}} e^{-x(1/2 - C/r^2A^2)} \expit(x)^2 \nonumber \\
    \sim & \frac{1}{\sqrt{2 \pi} rA} e^{-r^2 A^2/8} e^{C/2} \int_\mathbb{R} dx \frac{e^{-x/2}}{(1+e^{-x})^2}
    \sim \frac{\sqrt{\pi}}{4} \sqrt{\frac{w_2}{w_1}} \frac{1}{r \sqrt{1-s}} e^{\frac{-r^2(1-s)}{4}}
\end{align}
Recombining the different terms, this implies $f(s, \sigma) \sim \frac{\sqrt{\pi}}{2} \frac{\sqrt{w_1 w_2}}{r \sqrt{1-s}} e^{\frac{-r^2(1-s)}{4}}$

Let's now tackle $g$. Recall $\epsilon = \frac{1}{2R^2} \log \frac{w_*}{1-w_*}$.
\begin{align}
    g(m,\sigma) &= \mathbb{E}_{x \sim \mathcal{N}(0, 1)} \left[ 1 - 2\expit \left( 2 \sigma Rx + 2R^2m + \log \frac{w_*}{1-w_*} \right) \right] \nonumber \\
    &= 1 - 2 \mathbb{E}_{x \sim \mathcal{N}(0, 1)} \left[ \expit \left( 2 R^2 (\frac{\sigma}{R}x + m + \epsilon) \right) \right] \nonumber \\
    &= 1 - 2 \mathbb{E}_{x \sim \mathcal{N}(0, 1/r)} \left[ \expit \left( 2 R^2 (x + m + \epsilon) \right) \right] \nonumber \\&= 1 - 2 \expit(2R^2(m+\epsilon)) + \mathcal{O}(r^{-2})\\
\end{align}
Comparing the order of magnitude of $f$ and $g$, the low variance dynamics are dominated by $g$ which comes from the cross-entropy which attracts the means at $\pm \mu_*$.

\subsubsection{High variance limit ($\sigma \gg R$)}

We derive the high variance asymptotics of the entropic force $f$ and the cross-entropic force $g$. We denote $r = R/\sigma$ and consider $r \to 0$. Let's begin with $f$ and decompose it in the following way:
\begin{align}
    f(s, \sigma) &= \mathbb{E}_{x \sim \mathcal{N}(0, I_d)} \left[ w_1 \expit \left( r\sqrt{2(1-s)}x + \log \frac{w_2}{w_1}\right)^2 +  w_2 \expit \left( r\sqrt{2(1-s)}x + \log \frac{w_1}{w_2}\right)^2\right] \nonumber\\
    &+ \mathbb{E}_{x \sim \mathcal{N}(0, I_d)} \left[ w_1 \expit \left( r\sqrt{2(1-s)}x + r^2(s-1) + \log \frac{w_2}{w_1}\right)^2 \right. \nonumber \\
    & \left. \hspace{7em} +  w_2 \expit \left( r\sqrt{2(1-s)}x + r^2(s-1) + \log \frac{w_1}{w_2}\right)^2\right] \nonumber\\
    &- \mathbb{E}_{x \sim \mathcal{N}(0, I_d)} \left[ w_1 \expit \left( r\sqrt{2(1-s)}x + \log \frac{w_2}{w_1}\right)^2 +  w_2 \expit \left( r\sqrt{2(1-s)}x + \log \frac{w_1}{w_2}\right)^2\right]
\end{align}
For the first line, we make the change of variable $z = rx$. As $r \to 0$, we can use the asymptotic development of the convolution of a function with a gaussian of vanishing variance. As a result, the first line is equal to $w_1 \expit ( \frac{w_2}{w_1} )^2 + w_2 \expit ( \frac{w_1}{w_2} )^2 + \mathcal{O}(r^2) = w_1 w_2 + \mathcal{O}(r^2)$.

The difference between the second line and the third is bounded by Taylor equality at order $1$. Using $ \left| \expit \left( r\sqrt{2(1-s)}x + r^2(s-1) + \log \frac{w_2}{w_1}\right)^2 -  \expit \left( r\sqrt{2(1-s)}x + \log \frac{w_2}{w_1}\right)^2 \right| < ||(\expit^2)^\prime||_\infty r^2(s-1) < r^2$, we obtain that the difference is $\mathcal{O}(r^2)$.

As a results, $f(s, \sigma) = w_1 w_2 + \mathcal{O} \left( \frac{R^2}{\sigma^2} \right)$ as $\frac{R}{\sigma} \to 0$, uniformly in $s$.

Let's continue with $g$. Recall that $\epsilon = \frac{1}{2 R^2} \log \frac{w_*}{1-w_*}$ We decompose it in the following way:
\begin{align}
    g(m, \sigma) = &  \mathbb{E}_{x \sim \mathcal{N}(0, 1)} \left[ 1 - 2 \mathbbm{1}_{ \{ 2 \sigma Rx + 2R^2m + \log \frac{w_*}{1-w_*} > 0 \} } \right] \nonumber \\
     &+ \mathbb{E}_{x \sim \mathcal{N}(0, 1)} \left[ 1 - 2\expit \left( 2 \sigma Rx + 2R^2m + \log \frac{w_*}{1-w_*} \right) \right] - \mathbb{E}_{x \sim \mathcal{N}(0, 1)} \left[ 1 - 2 \mathbbm{1}_{ \{ 2 \sigma Rx + 2R^2m + \log \frac{w_*}{1-w_*} > 0 \} } \right] \nonumber \\
     = & \mathbb{E}_{x \sim \mathcal{N}(0, 1)} \left[ 1 - 2 \mathbbm{1}_{ \{x + r(m + \epsilon) > 0 \} } \right] \nonumber \\
     &+ \mathbb{E}_{x \sim \mathcal{N}(0, 1)} \left[ 1 - 2\expit \left( 2 \sigma Rx + 2R^2m + \log \frac{w_*}{1-w_*} \right) \right] - \mathbb{E}_{x \sim \mathcal{N}(0, 1)} \left[ 1 - 2 \mathbbm{1}_{ \{ x + r (m + \epsilon) > 0 \} } \right] \nonumber \\
\end{align}
The first term is straightforward and equal to $- \sqrt{\frac{2}{\pi}}r(m+\epsilon) + \mathcal{O}(r^3)$, by Taylor expanding the Gaussian density in $0$.
The second and the third term can be rewritten as:
\begin{align}
    &\mathbb{E}_{x \sim \mathcal{N}(0, 1)} \left[ 1 - 2\expit \left( 2 \sigma Rx + 2R^2m + \log \frac{w_*}{1-w_*} \right) \right] - \mathbb{E}_{x \sim \mathcal{N}(0, 1)} \left[ 1 - 2 \mathbbm{1}_{ \{ x + r (m + \epsilon) > 0 \} } \right] \nonumber \\
    =& 2 \mathbb{E}_{x \sim \mathcal{N}(0, 1)} \left[ - \expit \left( 2 \sigma Rx + 2R^2m + \log \frac{w_*}{1-w_*} \right) + \mathbbm{1}_{ \{ x + r (m + \epsilon) > 0 \} } \right] \nonumber \\
    =& 2 \mathbb{E}_{x \sim \mathcal{N}(2R^2(m + \epsilon), 2 \sigma R)} \left[ - \expit \left( x \right) + \mathbbm{1}_{ \{ x > 0 \} } \right] \nonumber \\
\end{align}
Note that the function $x \to - \expit \left( x \right) + \mathbbm{1}_{ \{ x > 0 \} }$ is odd an decays exponentially as $|x| \to \infty$. Thus, we can Taylor expand the Gaussian density as a function of $\frac{1}{2\sigma R}$ near $0$. Because even terms of this expansion are $0$, we get that the first non vanishing term is of order $\mathcal{O}(\frac{r}{\sigma^2})$.

Comparing the order of magnitude of $f$ and $g$, the high variance dynamics are dominated by $f$ which comes from the entropy and drive $\mu_1$ and $\mu_2$ away.

\section{Mode collapse probability estimate}
\label{app: sec: mode collapse probability estimate}

\subsection{Linearization around $(m_1=m_2=0, s=-1)$}

We study here the high temperature ($\beta \ll R^{-2}$) dynamical regime of the dynamics. At initialization, in high dimensions, $m_1 \sim \mathcal{N}(0, 1/d)$ and $m_2 \sim \mathcal{N}(0, 1/d)$ are independent.

From the above high variance limits of $f$ and $g$,
\begin{align}
    f(s,\sigma(t)) &= w_1 w_2 + \mathcal{O}(R^2/\sigma^2) \label{app: eq: f and g low beta asymptotic} \\
    g(m,\sigma(t)) &= - \sqrt{\frac{2}{\pi}} \frac{R}{\sigma}(m+\epsilon) + \mathcal{O}\left( R^2 / \sigma^2 \right).
\end{align}
We deduce that the high temperature dynamics are dominated by the entropic force $f$, driving the means apart and effectively projecting them onto the symmetric subspace $s=-1$ on a typical time scale $\mathcal{O}(w_1w_2) = \mathcal{O}(1)$. As $m_1$ and $m_2$ typical initial values scale as $1/\sqrt{d}$ (considering$\sqrt{d} \gg 1$), once $s \approx -1$, the dynamics can be understood by linearizing the coupled ODEs of \cref{eq: dynamical system annealing} around the point $(m_1=0, m_2=0, s=-1)$. The time evolution of the sum and difference of $m_1$ and $m_2$ decouple from $s$:
\begin{align}
    \dot{m_1} \! + \! \dot{m_2} =& - \left[ 2(1-\kappa)f(-1, \sigma(t)) + \frac{1}{2} g^\prime(0, \sigma(t)) \right] (m_1 \! + \! m_2) - g(0, \sigma(t)) \nonumber \\
    \dot{m_1} \! - \! \dot{m_2} =& - \frac{1}{2} g^\prime(0, \sigma(t)) (m_1 \! - \! m_2) \nonumber \\
    \dot{\delta s} =& - \left[ 4 \delta s (1-\kappa) f(-1, \sigma(t)) + (m_1 \! + \! m_2) g(0, \sigma(t)) \right].
\end{align}
where $\delta s = s+1$. We can explicitly solve the above linearized dynamical system for an arbitrary time dependant $\beta(t)$.
\begin{align}
    m_1(t) + m_2(t) &= \left( m_1(0) + m_2(0) \right) e^{-\int_0^t ds \, a(s)} - \int_0^t ds \, b(s) e^{-\int_s^t du \, a(u)} \\
    m_1(t) - m_2(t) &= \left( m_1(0) - m_2(0) \right) e^{-\int_0^t ds \, \frac{1}{2}g^\prime(0, \sigma(s))}
\end{align}
where $a(s) = 2(1-\kappa)f(-1, \sigma(s)) + \frac{1}{2} g^\prime(0, \sigma(s))$ and $b(s) = g(0, \sigma(s))$.

Assuming the high temperature limit, $a(s) = w_1 w_2 + \mathcal{O}(R/\sigma(s))$ and $b(s) = \mathcal{O}(R/\sigma(s))$. Hence, the sum $m_1(t) + m_2(t)$ is of order $\mathcal{O}(R/\sigma(t))$.

On the other hand, the difference $m_1 - m_2$ is unstable since $-g^\prime(0, \sigma(t)) = \sqrt{\frac{2}{\pi}} \frac{R}{\sigma(t)} + \mathcal{O}\left( R^2 / \sigma(t)^2 \right) > 0$. Assuming that the transient regime of the variance is over, we have $\beta(t) \sigma^2(t) = (1-\kappa)^{-1}$
\begin{align}
    m_1(t) - m_2(t) &= \left( m_1(0) - m_2(0) \right) e^{ \int_0^t \mathrm{d}s \, \sqrt{\frac{(1-\kappa)\beta(s) R^2}{2\pi}} } + \mathcal{O} \left( \beta R^2 \right).
\end{align}

\subsection{Mode collapse probability}

We derive the theoretical estimate of the mode collapse probability, assuming that the transition from the high-temperature regime to the low temperature is really abrupt and occurs at $\beta = \alpha R^{-2}$.

Given an annealing schedule $\beta(t)$, we denote $t_1$ such that $\beta(t_1) = \alpha R^{-2}$. In other words, we suppose that for $t<t_1$, the dynamics can be described by the high temperature limit $\beta \ll R^{-2}$, and that for $t>t_1$, the dynamics can be described by the low temperature limit. We stress that the transition at $\beta = \alpha R^{-2}$ is a modelling approximation: the crossover between regimes is smooth in the full ODE \cref{eq: dynamical system annealing}, while the sharp-transition formula makes the derivation tractable.

Leveraging the high temperature linearization of the dynamics until $t=t_1$, with a good approximation we have:
\begin{align}
    m_1(t_1) + m_2(t_1) &\ll 1 \\
    m_1(t_1) - m_2(t_1) &= \left( m_1(0) - m_2(0) \right) e^{ \int_0^{t_1} ds \, \sqrt{\frac{(1-\kappa)\beta(s) R^2}{2\pi}} }.
    \label{app: eq: at time t1}
\end{align}
For $t>t_1$, the dynamics is described by the low temperature limit, which means that the asymptotic fate of $m_1$ and $m_2$ are governed by the sign of $m_1(t_1) + \epsilon$ and $m_2(t_1) + \epsilon$ respectively. Mode collapse is avoided if $\text{min} (m_1(t_1), m_2(t_1)) < - \epsilon$ and $\text{max} (m_1(t_1), m_2(t_1)) > - \epsilon$. The previous condition is achieved if (recall \cref{app: eq: at time t1}):
\begin{align}
    |m_1(t_1) - m_2(t_1)|>2\epsilon \quad \Longleftrightarrow{} \quad |m_1(0) - m_2(0)|>2\epsilon e^{ -\int_0^{t_1} ds \, \sqrt{\frac{(1-\kappa)\beta(s) R^2}{2\pi}} }.
\end{align}
Since $m_1(0) - m_2(0)$ is the sum of two Gaussians of variance $1/d$ (in high dimensions), it is a Gaussian of variance $2/d$. Hence, we can compute the mode collapse probability $p(\beta)$:
\begin{align}
    p(\beta) = \int_{-2 \epsilon e^{-I(t_1)}}^{2 \epsilon e^{-I(t_1)}} \mathrm{d}x \sqrt{\frac{d}{4 \pi}} e^{-\frac{dx^2}{4}}
    \label{app: eq: mode collapse prob}
\end{align}
where we define: $$I(\beta) = \int_0^{t_1} \mathrm{d}s \, \sqrt{\frac{(1-\kappa)\beta(s) R^2}{2\pi}}.$$

\subsection{Focus on the exponential annealing schedule}
\label{app: sec: prob collapse for exponential schedule}

\begin{figure}[t]
\vskip 0.1in
\begin{center}
    \centerline{\includegraphics[width=\linewidth]{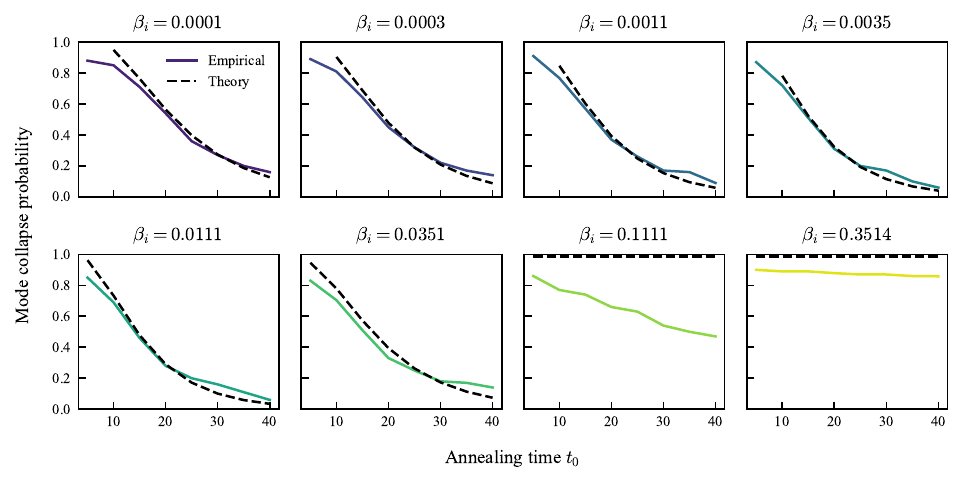}}
    \caption{\textit{Collapse probability for different initial temperatures, for Gaussian mixture student with an exponential annealing schedule.} Coloured curves are experimental mode collapse probability computed overs 100 random student mean initialization. The experimental data are the same as in \cref{fig:bimodal grid collapse prob and agreement}. The dashed black line are the theoretical mode collapse probability \cref{eq: prob collapse}, with $\alpha = 0.85$, and are only drawn for $\kappa<1$. Hyperparameters are $d=512$, $R=3$, and $w_*=0.8$.}
    \label{app: fig: collapse probability slices}
\end{center}
\end{figure}

Let's focus on the particular case of an exponential annealing schedule of the form:
\begin{align}
    \beta(t) = \min(\beta_i^{1-t/t_0}, 1).
\end{align}
Then,
\begin{align}
    \int_0^{t_1} {\rm d}s \sqrt{\beta(s)} = \sqrt{\beta_i} \int_0^{t_1} {\rm d}s e^{-\frac{s}{2t_0} \log \beta_i} = \frac{2 t_0}{\log 1/\beta_i} \left[ \frac{\sqrt{\alpha}}{R} - \sqrt{\beta_i} \right].
\end{align}
This yields
\begin{align}
    I(\beta_i, t_0) = \sqrt{\frac{2}{\pi}} \frac{\sqrt{(1-\kappa)}}{\kappa} \left[ \sqrt{\alpha} - \sqrt{R^2\beta_i} \right],
\end{align}
where $\kappa = \frac{\log(1/\beta_i)}{t_0}$ is the annealing rate. Reinjecting this integral into \cref{app: eq: mode collapse prob} we obtain a family of theoretical estimates of the mode collapse probability parametrized by $\alpha$. Since $\alpha/R^2$ corresponds to the end of the high temperature dynamical regime, we expect it to be of order one. \cref{app: fig: collapse probability slices} shows the agreement with the experimental curves when setting $\alpha \approx 0.85$. The agreement for high initial temperature (low $\beta_i$) is excellent. The agreement worsens as $\beta_i$ is increased to $1/R^2$, as expected since the high temperature dynamical regime description becomes less and less valid. When $\beta_i$ is already close to the crossover scale, the high-temperature phase is too short or absent, so $I(\beta_i, t_0)$ remains small even as $t_0$ changes. \Cref{eq: prob collapse} then predicts collapse probability close to one, explaining the black dashed line in the last panels.

The success of annealing depends on maximazing the integral $I(\beta_i, t_0)$. The dependence with respect to $t_0$ is increasing and intuitive: the longer the annealing time, the better. The integral has a unique global optimum with respect to $\beta_i$. In the regime $\kappa \ll 1$,
\begin{align}
    \frac{\partial I(\beta_i, t_0)}{\partial \beta_i} = 0 \Longleftrightarrow & \frac{1}{2}R - \frac{R \sqrt{\beta_i} - \sqrt{\alpha}}{\sqrt{\beta_i} \log \beta_i} = 0 \nonumber \\
    & \log (\beta_i/e) \frac{\beta_i}{e} = \frac{- \sqrt{\alpha}}{eR} \nonumber \\
    &\beta_i = e^2 W_0^2 \left( -\frac{\sqrt{\alpha}}{eR} \right)
\end{align}
where $W_0$ is the Lambert $W_0$ function defined by $W_0(z)e^{W_0(z)} = z$ On the real axis $W_0(x) < x$, and $W_0(x) \sim x$ as $x \to 0$.

Hence the optimal value of the initial temperature $\beta_i$ is such that $\beta_i < \alpha/R^2$ and $\beta_i \sim \alpha/R^2$ as $R \to \infty$.

We now focus on the limit where the initial temperature is high ($\beta_i \ll R^{-2}$) ---which is not optimal but relevant to practical cases when the true $R$ is a priori unknown. We have:
\begin{align}
    I(\beta_i, t_0) \sim \sqrt{\frac{2}{\pi}} \frac{ \sqrt{(1-\kappa)\alpha}}{\kappa}.
\end{align}
Asymptotically, the value of the integral only depends on the annealing rate $\kappa = \log(1/\beta_i)/t_0$. This is expected since the higher the annealing rate, the faster the temperature will go to $1$. Notably, it implies that if the initial temperature is high enough, the relevant quantity for the success of annealing is the annealing rate. \cref{fig: summary probability collapse} reports the probability of mode collapse with respect to the annealing rate. Increasing temperature curves rapidly collapse towards the asymptotic regime where mode collapse probability depends only on the annealing rate.


\section[Appendix]{Supplementary experimental results}

\subsection[Appendix]{Learnable weights}
\label{appendix: learnable weights}

Recall the minimal setting for the study of mode collapse in variational inference:
\begin{align}
    \pi(x) =& w_* \mathcal{N}(x|\mu_*, I_d) + (1-w_*)\mathcal{N}(x|-\mu_*, I_d) , \\
    q_\theta(x) =&  w_1 \mathcal{N}(x|\mu_1, \sigma_1^2I_d) + w_2\mathcal{N}(x|\mu_2, \sigma_2^2I_d).
\end{align}
To simplify the analytical treatment in the main body, we fix the student weights to be $w_1 = w_2 = 0.5$. In this section we provide experimental results when weights are learnable, that is to say $\theta = (\mu_1, \mu_2, \sigma_1, \sigma_2, w_1 , w_2)$ with constrain $w_1+w_2=1$.

\begin{figure}[t]
    \centering
    \begin{subfigure}{0.50\linewidth}
        \centering
        \includegraphics[width=\linewidth]{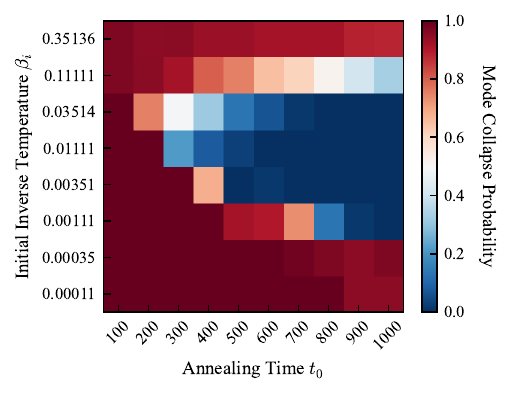}
        \caption{}
        \label{fig: learnable weights prob collapse}
    \end{subfigure}
    \hfill
    \begin{subfigure}{0.49\linewidth}
        \centering
        \includegraphics[width=\linewidth]{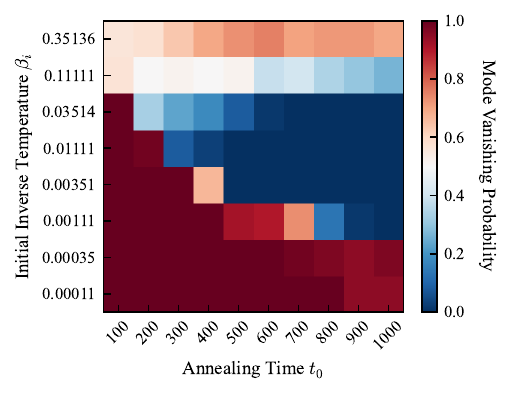}
        \caption{}
        \label{fig: learnable weights vanishing prob}
    \end{subfigure}
    \caption{\textit{Annealed VI mode collapse probability with Gaussian mixture student with learnable weights.} Both the student and the teacher are bimodal gaussian mixtures. The form of the annealing scheme is fixed by \cref{exponential annealing schedule} and different values of the initial temperature $\beta_i$ and annealing time $t_0$ are explored. Hyperparamters are $d=512$, $R=3$ and $w_*=0.8$. For each values of $(\beta_i, t_0)$, the mode collapse probability is obtained using 50 different random student mean initialization (student means are initialized randomly on the sphere of radius $R$, variances are initialized at $1/\beta_i$ and weights at $0.5$). We use Adam optimizer with learning rates $\eta_\mu = 0.003$ for the means, $\eta_{\sigma^2} = 0.02 \beta^{-1}$ for the variances and $\eta_w = 0.01$ for the weights.\\
    \cref{fig: learnable weights prob collapse} reports the probability of mode collapse (due to mean alignment or mode vanishing), \cref{fig: learnable weights vanishing prob} reports the probability mode vanishing.}
\end{figure}

Learnable weights introduce a supplementary failure mode, as mode collapse can also be driven by vanishing weights (\citep{soletskyi_theoretical_2024}). Weight vanishing happens when either of one of the 2 student weights irreversibly converge to $0$, meaning that the student distribution becomes effectively unimodal. \Cref{fig: learnable weights prob collapse} shows the mode collapse probability over a range of initial inverse temperature $\beta_i$ and annealing time $t_0$ ($t_0$ in Adam iterations). In this figure, mode collapse due to both vanishing weights and mean alignment ($\mu_1$ and $\mu_2$ converging to the same target mode) are taken into account. Compared to the fixed weight case (\cref{fig:bimodal grid collapse prob and agreement}), we observe that the model has the same qualitative behavior, with a more abrupt transition between the success or failure of annealing.

The right panel \cref{fig: learnable weights vanishing prob} shows the probability of observing vanishing weights. We observe that the low $t_0$ and low $\beta_i$ region is prone to mode vanishing. In fact, comparing with \cref{fig: learnable weights prob collapse}, we deduce that mode collapse in this parameter region is mostly due to vanishing weights. We stress that the vanishing weight probability undergoes the same tradeoff identified in this work: the higher the initial temperature, the longer the annealing time has to be in order to prevent it.

\subsection{RealNVP training}
\label{appendix: realnvp training}

The architecture used for normalizing flows are RealNVP \citep{dinh_density_2017}, with 8 affine coupling layers. In each layer, the neural networks parametrizing the scale and the translation are MLPs with depth 4 and hidden dimension $h=4d$, if $d$ is the latent space dimension. We use Adam optimizer with learning rate $\eta = 0.002$. Our implementation is provided in the code made available.

\subsection{5-mode Gaussian mixture target}
\label{appendix: 5 mode gaussian target}

This section provides details about the 5-mode Gaussian mixture target distribution used in \cref{fig: realnvp grid prob collapse} (right panel). We generated a 5-mode target following this procedure. First, we drew the 5 mode locations uniformly on the sphere of radius $R=3\sqrt{2}$. Because those 5 modes will be approximately orthogonal in high dimensions, this ensures that the distance between any pair of target mode location is $\approx 6$, consistent with the bimodal target used in \cref{fig: realnvp grid prob collapse}. Then, one of the modes was assigned probability $0.5$ while the remaining 4 were assigned probability $0.125$.

\subsection{Dependence on the annealing rate}
\label{appendix: dependence on annealing rate}

\begin{figure*}[h!]
    \begin{center}
    \centerline{\includegraphics[width=\linewidth]{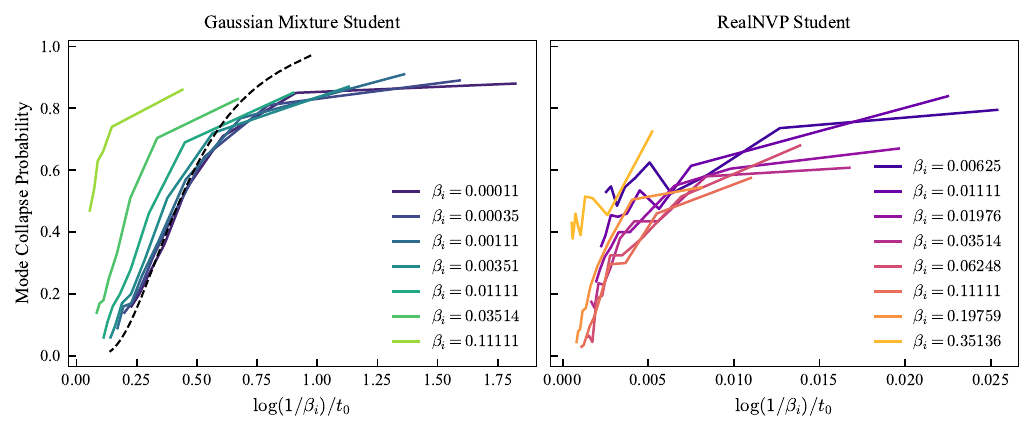}}
    \caption{Probability of mode collapse as a function of the annealing rate $\kappa = \log(1/\beta_i)/t_0$, for an exponential annealing schedule given by \cref{exponential annealing schedule}. For the Gaussian mixture student the data are the same as \cref{fig:bimodal grid collapse prob and agreement}, hyperparameters are $d=512, R=3, w_*=0.8$. The dashed black line is the high initial temperature ($\beta_i \to 0$) asymptote of the mode collapse probability \cref{eq: prob collapse} ($\alpha = 0.85$), only drawn for $\kappa<1$. For the realNVP student the data are the same as \cref{fig: realnvp grid prob collapse} (left), hyperparameters are $d=128, R=3, w_*=0.8$. For the Gaussian mixture student $t_0$ is in gradient-flow time units, for the RealNVP student in Adam iterations, so the values of $\kappa$ are not comparable between the two panels.}
    \label{fig: summary probability collapse}
    \end{center}
\end{figure*}

\begin{figure}[h!]
    \centering
    \includegraphics[width=0.5\linewidth]{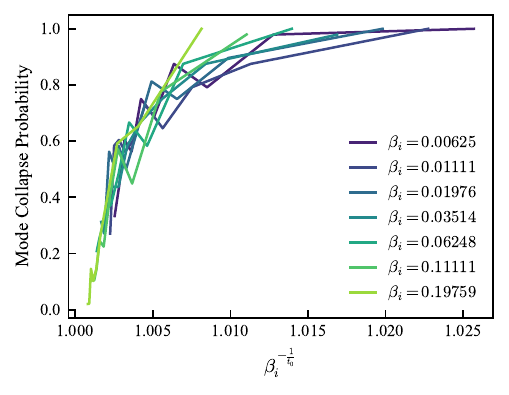}
    \caption{Annealed VI mode collapse probability with RealNVP student on a 5-mode target as a function of $e^\kappa$. Experimental setting and data are the same as in \cref{fig: realnvp grid prob collapse} (right).}
    \label{fig: prob scaling complex target}
\end{figure}

\cref{fig: summary probability collapse} shows that the experimental mode collapse probabilities as a function of the annealing rate collapse to a single curve as $\beta_i \rightarrow 0$. This confirms that for sufficiently high initial temperatures, the mode collapse probability depends essentially on the annealing rate alone. The dotted black line on the left column is the theoretical ($\beta_i \rightarrow 0$) asymptote of the mode collapse probability \cref{eq: prob collapse}. The right panel recovers for a complex generative model architecture the behavior predicted by our theoretical analysis, which further confirms the relevance of our theoretical insights in practical settings. Furthermore, \cref{fig: prob scaling complex target} extends this results to a more complex target (5-mode Gaussian mixture).

\end{document}